\documentclass{article}

\usepackage[final, nonatbib]{neurips_2021_data}
\usepackage[numbers]{natbib}

\usepackage[utf8]{inputenc} \usepackage[T1]{fontenc}    \usepackage{hyperref}       \usepackage{url}            \usepackage{booktabs}       \usepackage{amsfonts}       \usepackage{nicefrac}       \usepackage{microtype}      \usepackage[dvipsnames]{xcolor}         \usepackage{graphicx}
\usepackage{makecell}
\usepackage{multirow}
\usepackage{caption} 
\usepackage{tablefootnote}
\usepackage{arydshln}
\usepackage{setspace} 
\usepackage[utf8]{inputenc} \usepackage[T1]{fontenc}    \usepackage{hyperref}       \usepackage{url}            \usepackage{booktabs}       \usepackage{amsfonts}       \usepackage{nicefrac}       \usepackage{microtype}      \usepackage{xcolor}         \usepackage{multirow}
\usepackage{multicol}
\usepackage{makecell}
\usepackage{placeins}
\usepackage{graphicx}

\usepackage{subcaption}

\usepackage[utf8]{inputenc} \usepackage[T1]{fontenc}    \usepackage{hyperref}       \usepackage{url}            \usepackage{booktabs}       \usepackage{amsfonts}       \usepackage{nicefrac}       \usepackage{microtype}      \usepackage{xcolor}         \usepackage{graphicx}
\usepackage{makecell}
\usepackage{multirow}
\usepackage{caption} 
\usepackage{moreverb}
\usepackage{listings}
\usepackage{amsmath}
\usepackage{caption} 
\usepackage{tablefootnote}

\usepackage{listings}
\usepackage{xcolor}

\definecolor{codegreen}{rgb}{0,0.6,0}
\definecolor{codegray}{rgb}{0.5,0.5,0.5}
\definecolor{codepurple}{rgb}{0.58,0,0.82}
\definecolor{backcolour}{rgb}{0.98,0.98,0.98}

\lstdefinestyle{mystyle}{
    backgroundcolor=\color{backcolour},   
    commentstyle=\color{codegreen},
    keywordstyle=\color{magenta},
    numberstyle=\tiny\color{codegray},
    stringstyle=\color{codepurple},
    basicstyle=\ttfamily\footnotesize,
    breakatwhitespace=false,         
    breaklines=true,                 
    captionpos=b,                    
    keepspaces=true,                 
    numbers=left,                    
    numbersep=5pt,                  
    showspaces=false,                
    showstringspaces=false,
    showtabs=false,                  
    tabsize=2
}

\title{HiRID-ICU-Benchmark~--- A Comprehensive Machine Learning Benchmark on High-resolution ICU Data}
\author{
    Hugo Yèche $^{1}$ \thanks{Equal contribution} \And Rita Kuznetsova $^{1}$ \footnotemark[1] \And  Marc Zimmermann $^{1}$ \AND Matthias Hüser $^{1}$ \And Xinrui Lyu $^{1}$ \And  Martin Faltys $^{1,2}$ \And Gunnar Rätsch $^{1}$ \And
    \texttt{\{hyeche,mkuznetsova,marczim,mhueser,xlyu,mfaltys,raetsch\}@inf.ethz.ch}\\
    
    \small{$^{1}$Department of Computer Science, ETH Zürich }\\
    \small{ $^{2}$Department of Intensive Care Medicine, University Hospital, and University of Bern}\\
}

\begin{document}

\maketitle

\begin{abstract}
The recent success of machine learning methods applied to time series collected from Intensive Care Units (ICU) exposes the lack of standardized machine learning benchmarks for developing and comparing such methods. While raw datasets, such as MIMIC-IV or eICU, can be freely accessed on Physionet, the choice of tasks and pre-processing is often chosen ad-hoc for each publication, limiting comparability across publications. In this work, we aim to improve this situation by providing a  benchmark covering a large spectrum of ICU-related tasks. Using the HiRID dataset, we define multiple clinically relevant tasks in collaboration with clinicians.  In addition, we provide a reproducible end-to-end pipeline to construct both data and labels. Finally, we provide an in-depth analysis of current state-of-the-art sequence modeling methods, highlighting some limitations of deep learning approaches for this type of data. With this benchmark, we hope to give the research community the possibility of a fair comparison of their work.

\smallskip
\small
\textbf{Software Repository: }\url{https://github.com/ratschlab/HIRID-ICU-Benchmark/}
\end{abstract} \section{Introduction}\label{intro}
Severely ill patients require treatment and surveillance in Intensive Care Units (ICU). Critical health conditions are characterized by the presence or risk of developing life-threatening organ dysfunction. During a patient's stay in the ICU, continuous monitoring of organs function parameters enables early recognition of physiological deterioration and rapid commencement of appropriate interventions. Recent research shows the great success of machine learning methods when applied to ICU time series~\cite{ncle, horn2020set}. One of the main goals of previous works was to develop new methods for prediction tasks relevant to clinical decision-making. Exemplary of such tasks are alarm systems that predict different types of organ failure~\cite{hyland2020early,tomavsev2019clinically}.

To develop and evaluate such methods only a small number of large-scale ICU datasets are freely-accessible: The MIMIC-III~\cite{johnson2016mimic} and IV~\cite{MIMIC-IV} datasets, AmsterdamUMCdb~\cite{ams}, HiRID \citep{hirid} and the eICU Collaborative Research Database~\cite{pollard2018eicu}. However, these datasets are not provided in a pre-processed form directly suitable for machine learning nor do they have well-defined tasks, making it impossible to fairly compare works~\cite{johnson2017reproducibility}. While some pre-processed alternatives with well-defined tasks exist \citep{goldberger2000physiobank, reyna2019early}, they are often lacking in terms of size and diversity of tasks. We provide more details about this in section~\ref{rel_work}. This leads to situations where works compare methods on their private data \citep{tomavsev2019clinically} or only on limited data and number of tasks. Also the lack of relevant clinical sub-tasks for benchmarking hinders the development of new methods for clinical decision support systems~\cite{harutyunyan2019multitask}. Finally, as in other fields, in recent datasets such as HiRID \citep{hirid} the time resolution of data has greatly increased. However, no benchmark on ICU time series using such high-resolution datasets currently exists.

To improve this situation, in this paper we provide an in-depth benchmark based on the HiRID dataset~\citep{hirid, hyland2020early}\footnote{\url{https://physionet.org/content/hirid/1.1.1/}}, which was released on Physionet~\cite{hirid} alongside the publication on the circulatory Early Warning Score (circEWS)~\cite{hyland2020early}. HiRID is a freely accessible critical care dataset containing data recorded at the Department of Intensive Care Medicine, the University Hospital of Bern, Switzerland (Inselspital). The dataset was developed in cooperation with the Swiss Federal Institute of Technology (ETH), Zürich, Switzerland. We define a new benchmark on HiRID composed of various clinically relevant tasks and provide a comprehensive pipeline, which includes all steps from preprocessing to model evaluation. To assess different aspects of the benchmarked machine learning methods, we diversify the tasks around specific challenges of ICU data such as prediction frequency, class imbalance, or organ dependency of the task. To profit from data acquisition advances and allow improvement on longer time series, we use a resampled data resolution of 5 min. HiRID has a higher time resolution than any other published critical care dataset and it motivates us to provide a comprehensive benchmark suite on this dataset. Also, we believe that this dataset will facilitate the construction of new predictive methods for the healthcare field, going beyond ICU time series.

The main contributions of this paper are:
\begin{itemize}
    \item We developed a comprehensive, end-to-end pipeline for time-series analysis of critical care data based on the recently published HiRID dataset. This pipeline includes the following stages: data preprocessing mode, training mode, and evaluation mode. 
    \item We proposed and implemented a variety of tasks relevant to healthcare workers in the ICU, diversified in terms of type, prediction resolution, and label prevalence. The tasks cover all major organ systems as well as the general patient state. We included both regression and classification (binary and multi-class) tasks. 
    \item By providing a comprehensive benchmark on a set of canonical tasks, we give the research community around predictive modeling on  ICU time series the possibility for the clear comparison of their methods.
\end{itemize}
The paper is organized as follows: in Section~\ref{rel_work} we provide an overview of existing ICU datasets and benchmarking papers. We provide details about the HiRID dataset and introduce the tasks defined in collaboration with clinicians in Section~\ref{bench_design} and give more details on the tasks in \textsc{Appendix A: Dataset Details}. Section~\ref{pipeline} illustrates the pipeline design, with more details given in \textsc{Appendix B: HiRID-ICU-Pipeline Details}. Section~\ref{exp} describes the experiment and ablation study. In Section~\ref{disc} we discuss the observed results and relate this paper to other benchmarks and related tasks relevant for clinicians. 

 \section{Related Work}\label{rel_work}
The main goal of this work is to provide a benchmark on the HiRID dataset for various clinical prediction tasks of interest. We describe here other ICU datasets as well as existing benchmarks for ICU data.
\paragraph{ICU time-series datasets} 
There are several widely-used, freely-accessible datasets consisting of ICU time series. MIMIC-III~\cite{johnson2016mimic} is the oldest and most widely used ICU dataset. It consists of physiological measurements as well as information about laboratory tests. Physiological measurements are recorded with a maximum resolution of 1 hour. The results of laboratory tests are collected at irregular time intervals. Moreover, there are static features like gender, age, diagnosis, etc. available. The dataset consists of information recorded about 40,000 ICU stays at Beth Israel Deaconess Medical Center (BIDMC), Boston, MA, USA. The median of the patient stay length is 2 days. The eICU Collaborative Research Database~\cite{pollard2018eicu} is a large multicenter critical care database made available by Philips Healthcare in partnership with the MIT Laboratory for Computational Physiology. It contains data associated with over 200,000 patient stays, but the public version does not reach the granularity of other datasets in terms of time resolution and data elements.
The first version of AmsterdamUMCdb~\cite{ams} was released in November 2019. Its current version from March 2020 contains data related to 23,172 ICU and high dependency unit admissions of adult patients from 2003 - 2016 from Amsterdam University Medical Centers. The data includes clinical observations like vital signs, clinical scores, device data, and lab results.
\paragraph{Benchmarks on ICU time-series.} Among works using the openly available datasets mentioned above, to the best of our knowledge, only a single standardized benchmark exists, MIMIC-III benchmark by Harutyunyan et al.~\cite{harutyunyan2019multitask}. In that work four tasks were proposed, two requiring a single prediction per patient stay and two dynamic tasks with more frequent prediction, one per hour. In addition, while not proposing a benchmark, Jarrett et al.~\cite{jarrettclairvoyance} developed a standardized pipeline for medical time series, called Clairvoyance. They also provided results on several datasets, including MIMIC. In this spirit, some packages address a specific family of tasks, for example, classification~\cite{faouzi2020pyts} and forecasting~\cite{guecioueurpysf}. Finally, some public challenges, with curated data, were proposed in the past, e.g.~the early prediction of sepsis (Physionet 2019 challenge~\cite{reyna2019early}) or mortality prediction (Physionet 2012 challenge \cite{citi2012physionet}). However, the provided datasets are smaller than HiRID and are built around a single task.

\section{Benchmark Design}\label{bench_design}

\subsection{The HiRID Dataset}

HiRID~\citep{hirid, hyland2020early} is a freely accessible critical care dataset containing data from more than 33,000 patient admissions to the Department of Intensive Care Medicine, the University Hospital of Bern, Switzerland (Inselspital) from January 2008 to June 2016. It was released on Physionet~\cite{hirid} alongside the publication of the circulatory Early Warning Score (circEWS)~\cite{hyland2020early}. It contains de-identified demographic information and a total of 712 routinely collected physiological variables, diagnostic test results, and treatment parameters. HiRID has a higher time resolution than any other published ICU dataset, particularly for bedside monitoring, with most vital signs recorded every 2 minutes, which motivates us to provide a comprehensive benchmark suite on this dataset. Demographic information
about the patient cohort are displayed in Appendix Table 1.
\begingroup
\setlength{\tabcolsep}{3pt}
\renewcommand*{\arraystretch}{1}
\begin{table}[!hb]
  \footnotesize    
  \centering
  \caption{{Definition of prediction tasks contained
         in the HiRID-ICU benchmark suite}}
  \begin{tabular}{lll}
    \toprule
    \textbf{Task name}     & \textbf{Task type}     & \textbf{Task description} \\ \midrule
    ICU mortality & \makecell[l]{Binary classification, \\ one prediction per stay}  & \makecell[l]{Predicted at 24h after \\ admission to the ICU.}     \\ \hdashline[0.3pt/1pt]
    Patient phenotyping & \makecell[l]{Multi-class classification, \\ one prediction per stay} & \makecell[l]{Classifying the patient after 24h\\ regarding the  admission diagnosis,\\ using the APACHE group II and IV labels\tablefootnote{APACHE II and IV \cite{Zimmerman2006-of, Knaus1985-iw} are subsequent versions of the major illness severity score used in the ICU. They also introduce a patient grouping according to admission reason. We use an aggregate of these two groupings for this task (see \textsc{Appendix A: Dataset Details})} } \\
    \midrule
    Circulatory failure \tablefootnote{Circulatory failure is defined as Lactate $>2$mmol/l and either mean arterial blood pressure $<65$mmHg or administration of any vasoactive drug.} & \makecell[l]{Binary classification,\\ dynamic prediction throughout stay} & \makecell[l]{Continuous prediction of onset of \\ circulatory failure in the next 12h, \\given the patient is not in failure now.}     \\\hdashline[0.3pt/1pt]
    
    Respiratory failure\tablefootnote{Respiratory failure is defined according to the Berlin definition \citep{ARDS_Definition_Task_Force2012-hl} as a P/F ratio $< 300$ mmHg.} & \makecell[l]{Binary classification,\\ dynamic prediction throughout stay} & \makecell[l]{Continuous prediction of onset\\ of respiratory failure in the next 12h, \\ given the
    patient is not in failure now.}  
    \\
    \midrule
    Kidney function & \makecell[l]{Regression,\\ dynamic prediction throughout stay} & \makecell[l]{Continuous prediction of urine production \\ in the next 2h as an average rate in ml/kg/h. \\
    The task is predicted at irregular intervals.}  \\\hdashline[0.3pt/1pt]

    Remaining length of stay & \makecell[l]{Regression, \\dynamic prediction throughout stay} & \makecell[l]{Continuous prediction of the \\ remaining ICU stay duration.}      \\
    \bottomrule
  \end{tabular}\label{tab:tasks}
\end{table}
\endgroup

\subsection{Prediction Tasks}\label{tasks}

Our benchmark suite focuses on clinically relevant prediction tasks with a large diversity in the machine learning task types. From a clinical point of view, the tasks cover most major organ systems as well as the general patient state. The major organ systems include the cardiovascular, kidney, and respiratory systems. For each organ system, we provide a prediction task related to the main organ function.
Length of stay, mortality, and patient phenotyping are chosen to assess an overall patient state.
From a machine learning point of view, our suite contains regression and classification (binary and multi-class) tasks. We included tasks with different degrees of class imbalance to diversify the spectrum further and enable the comparison of methods on e.g. highly imbalanced tasks. We chose tasks performed online throughout the stay (every 5 minutes) and at fixed time-points of the stay, such
as 24h after ICU admission, which capture a more long-term state of the patient.
To enhance reproducibility, we include two tasks previously considered in \cite{harutyunyan2019multitask}, mortality, and remaining length-of-stay prediction. Table~\ref{tab:tasks} contains the full list of task and their detailed descriptions.

\section{Pipeline Design}\label{pipeline}
\begin{figure}[h]
    \centering
    \includegraphics[scale=0.39]{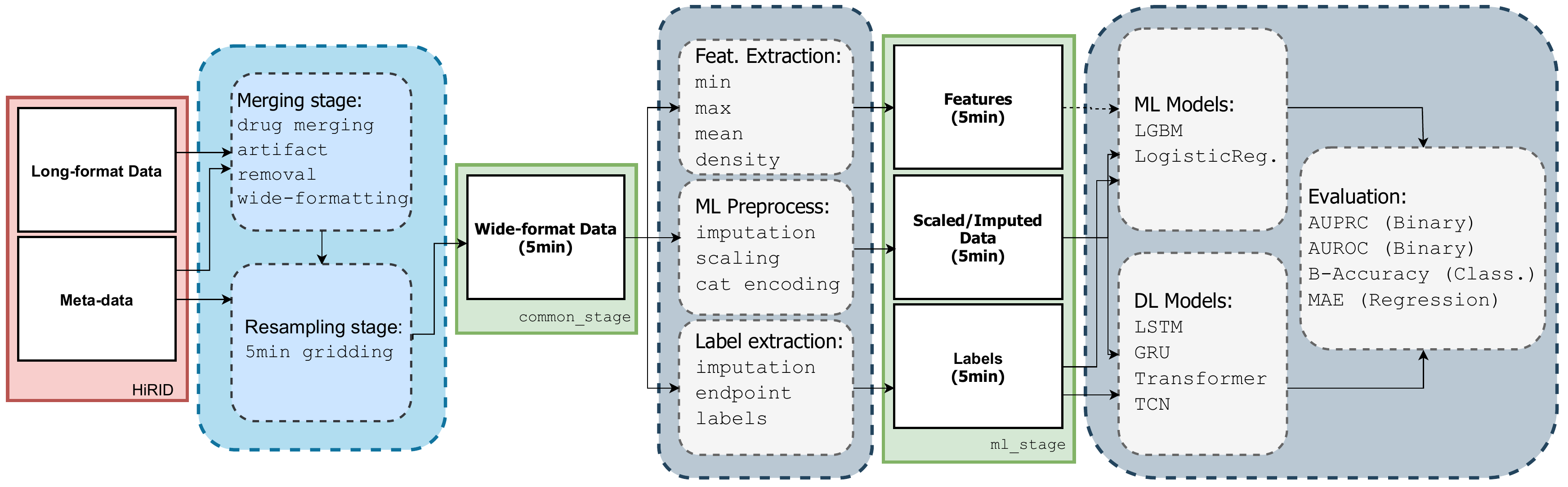}
    \caption{\textit{Detailed Pipeline.} (Red) Raw Long-format Data.  (Green) Wide-format data. (Blue) Common data pre-processing.  (Grey) Modeling depending stages.}
    \label{fig:pipeline-detailed}
\end{figure}

Figure~\ref{fig:pipeline-detailed} shows an overview of the major HiRID-ICU pipeline steps. The pipeline is designed using the \textit{preprocess-train-predict} paradigm. We provide more details about it in \textsc{Appendix B: HiRID-ICU Pipeline Details}  and the README section of the software repository\footnote{\url{https://github.com/ratschlab/HIRID-ICU-Benchmark}}. The preprocessed data contains two versions, \texttt{common\_stage} and \texttt{ml\_stage}. The former is independent of modeling choices and serves as the starting point for future works with custom pre-processing choices. The latter is a compatible version for our pipeline with our categorical encoding, imputation, and scaling choices.

\subsection{Data Pre-processing}
In its public version, HiRID, as any real-world dataset, contains certain artifacts that require pre-processing. As pointed out by \citep{bellamy2020evaluating} for MIMIC-III, individual pre-processing in each work avoids a fair comparison of them. To this effect, we aim to provide a modular and reproducible pipeline.
Patient EHRs in HiRID are stored in a long table format where each row of the table is a record containing the measurement value of a specific variable at a specific time for a patient, which cannot be used as a ready input for machine learning models.

\paragraph{Wide-format Merging}
To obtain a more compact format, the first pre-processing step in our pipeline is to transform the long table of patient EHRs into feature matrices, where each column represents a clinical concept, which we call the wide-format. Such a data format represents an irregularly sampled multivariate time series. At this step, we also remove any physiologically impossible measurement.

\paragraph{High-Resolution Gridding} After this merging step, we further compact the dataset by re-sampling it to a 5 minute resolution. Thus, each time step contains the last value measured in the last 5 min if it exists, or left empty otherwise. This gridding strategy is similar to the one used by \citep{hyland2020early}. We refer to the output of this step as the \texttt{common\_stage} in Fig.\ref{fig:pipeline-detailed}. Because it is independent of modeling choices, this stage provides a starting point for future approaches using different imputation and scaling choices.

\paragraph{Processing for Machine Learning} In the second part of the pipeline, we process the common stage of the data to be compatible with ML models' expected input format. For this, we first use forward-filling imputation for each stay. Then, we apply one-hot encoding for categorical variables and scale the remaining ordinal or continuous variables. We standard scaled all variables with the exception of the time since admission and admission date, which we min-max scaled. By doing scaling globally, we ensure to preserve patients' specificity (e.g.: tachycardia). We refer to the output of this stage as the \texttt{ml\_stage} as it is dependent on our modeling choices.

\subsection{Hand-engineered Feature Extraction}
In the original paper describing the HiRID dataset~\citep{hyland2020early}, the authors showed that boosted tree ensembles such as LGBM \citep{ke2017lightgbm}, when provided with hand-engineered features, outperform state-of-the-art deep learning methods. Based on this observation, we include in our pipeline the possibility to extract such features from the \texttt{common\_stage} of the data. For our models, we extracted four features for each non-categorical variable over the entire history: \textit{minimum past value}, \textit{maximum past value}, \textit{mean past value} and \textit{density of measurement}, which is the proportion of time points where a value is provided among all possible time points in the history\footnote{This is done on the regularly sampled version of the data}. These features are then included in the \texttt{ml\_stage}.

\subsection{Label Construction and Splitting}
 We construct prediction task labels using the provided measurements and meta-data for both continuous and stay-level tasks. As an intermediate step for label construction, we use a forward imputed version of the data, as in the modeling stage. Concerning the experimental design, we use a random split of patients. The training set contains 70\% of the patients and validation and test sets each contain 15\%. The temporal splitting strategy as used by Hyland et al.~\citep{hyland2020early} would be more clinically relevant but information about admission time was removed to preserve anonymity when the dataset was originally published. While longer stays exist in the dataset, for computational reasons, we limited labeling to the first 7 days of stays (2016 steps). This cropping affects less than 6\% of all stays.

\begin{table}[ht!]
\footnotesize
\caption{Label statistics for each of the tasks, in the training, validation and test sets. As a metric, for binary classification tasks, the positive label prevalence is reported. For multi-class classification tasks, the class prevalence of the minority class is reported. Finally, for regression tasks the median of the label 
distribution is reported. In parentheses the number
of samples is reported. M: Million.}
\begin{center}
\begin{tabular}{llll}
\toprule
\textbf{Task name} & \textbf{Train set} & 
\textbf{Validation set} & \textbf{Test set} \\
\midrule
Circ. failure & 4.3 \% (n=14.12M) & 4.1 \% (n=3.01M) & 4.1 \% (n=2.96M) \\
Resp. failure & 38.3 \% (n=5.58M) & 37.6 \% (n=1.21M) & 37.4 \% (n=1.20M) \\
\midrule
Mortality & 8.7 \% (n=10525) & 7.1 \% (n=2206) & 8.3 \% (n=2231) \\
Phenotyping & 0.2 \% (n=10470) & 0.1 \% (n=2194) & 0.1 \% (n=2217) \\
\midrule
Kidney function & 1.17 ml/kg/h (n=341424) & 1.12 ml/kg/h (n=71549) & 1.18 ml/kg/h (n=70642) \\
Rem. LOS & 41.04h (n=15.15M) & 41.51h (n=3.22M) & 39.64h (n=3.17M) \\
\bottomrule
\end{tabular}\end{center}
\label{tab:label-stats-per-split}
\end{table}

\subsection{Model Training and Evaluation}
The final part of the pipeline contains an end-to-end machine learning suite to train and evaluate our models, depicted on the right hand side of Fig.\ref{fig:pipeline-detailed}.Machine learning (ML) approaches were implemented using \texttt{scikit-learn}\cite{scikit-learn} and \texttt{lightgbm}\cite{ke2017lightgbm}, whereas deep learning (DL) approaches were implemented in \texttt{pytorch} \cite{paszke2019pytorch}. All DL models were trained using Adam optimizer \cite{kingma2014adam}, with a cross-entropy objective for classification tasks and mean-squared error (MSE) for regression tasks. For classification we provide the possibility to balance loss weights according to class prevalence as in \cite{king2001logistic}.

For the evaluation of models, we use a range of metrics relevant to each task. For classification tasks, we considered AUROC\footnote{\url{https://scikit-learn.org/stable/modules/generated/sklearn.metrics.roc\_auc\_score.html}} and AUPRC\footnote{\url{https://scikit-learn.org/stable/modules/generated/sklearn.metrics.average\_precision\_score.html}} metrics in the binary case, and balanced accuracy (B-Accuracy) \cite{brodersen2010balanced} in the multi-class one. For regression tasks, we used mean absolute error (MAE) as a comparison metric. Regardless of the task or model, we used the \texttt{scikit-learn} implementation for all metrics. More details about this stage of the pipeline can be found in \textsc{Appendix B: HiRID-ICU-Pipeline Details}.

   \section{Experiments}\label{exp}

\subsection{Settings}

For all models, we tuned specific hyper-parameters using random search. Each randomly picked set of parameters was run with 3 different random initializations. We then selected hyper-parameters on the validation set performance for either AUPRC, B-Accuracy, or MAE. All models were trained with early stopping on the validation loss. Further details about hyper-parameters can be found in \textsc{Appendix B: HiRID-ICU-Pipeline Details}.

Because of the class imbalance existing in classification tasks, we considered balanced loss weights for all methods. However as further discussed in subsection \ref{ablation:weight}, this technique was relevant only for the Patient Phenotyping task. For regression tasks, we min-max scaled the labels at training time to avoid exploding gradients.

\subsection{Benchmarked Methods}\label{ml_models}

In our proposed benchmark, we considered two groups of machine learning algorithms. The first group consists of regular machine learning algorithms, which as shown are highly effective for ICU-related tasks \citep{sadeghi2018early,harutyunyan2019multitask,hyland2020early}. It is composed of a Gradient Boosting method with LightGBM \citep{ke2017lightgbm} and Logistic Regression. The second group is focused on deep learning methods. We select the most commonly used sequence models for this group: Recurrent neural networks (LSTM~\citep{hochreiter1997long} and GRU~\citep{cho2014learning}), convolutional neural networks (CNN), in particular, temporal convolutional networks (TCN)~\cite{bai2018empirical} and Transformer models~\cite{vaswani2017attention}.

\subsection{Benchmarking Models on High-resolution ICU Data}
In this section, we compare the previously described methods on all tasks. While DL approaches are provided with the entire history for all time points, ML methods use only the values of the current step as an input. Thus one would expect the latter models to perform significantly worse due to the lower amount of information provided.
\begin{table}[!ht]
    \footnotesize
    \centering
    \caption{\textit{Benchmark of methods for stay level tasks}.(Top rows) ML methods; (Bottom rows) DL methods. All scores are averaged over 10 runs with different random seeds such that the reported score is of the form $mean \pm std$. In bold are the methods within one standard deviation of the best one. Classification metrics were scaled to 100 for readability purposes.}
\begin{tabular}{l|cc|c}
\toprule
Task & \multicolumn{2}{c|}{ICU Mortality} &     Patient Phenotyping  \\
\midrule
Metric & AUPRC ($\uparrow$) & AUROC ($\uparrow$) & B-Accuracy ($\uparrow$) \\
\midrule
LR                 &     58.1 $\pm$ 0.0 &     89.0 $\pm$ 0.0 &          39.1 $\pm$ 0.0\\
LGBM               &     54.6 $\pm$ 0.8 &     88.8 $\pm$ 0.2 &          40.4 $\pm$ 0.8  \\
LGBM w. Feat.      &     \textbf{62.6} $\pm$ 0.0 &     90.5 $\pm$ 0.0 &          \textbf{45.8} $\pm$ 2.0  \\
\midrule
\midrule
GRU                &     60.3 $\pm$ 1.6 &     90.0 $\pm$ 0.4 &          39.2 $\pm$ 2.1 \\
LSTM               &     60.0 $\pm$ 0.9 &     90.3 $\pm$ 0.2 &          39.5 $\pm$ 1.2  \\
TCN                &     60.2 $\pm$ 1.1 &     89.7 $\pm$ 0.4 &          41.6 $\pm$ 2.3  \\
Transformer        &     61.0 $\pm$ 0.8 &     \textbf{90.8} $\pm$ 0.2 &          42.7 $\pm$ 1.4 \\
\bottomrule
\end{tabular}

    \label{tab:stay-level}
\end{table}

\begin{table}[!ht]
    \centering
    \footnotesize
\setlength\tabcolsep{4pt}
\caption{\textit{Benchmark of methods for online monitoring tasks}. (Top rows) ML methods; (Bottom rows) DL methods. All scores are averaged over 10 runs with different random seeds such that the reported score is of the form $mean \pm std$. In bold are the methods within one standard deviation of the best one. Classification metrics were scaled to 100 for readability purposes. MAE is in units ml/kg/h for Kidney Function and in hours for Remaining LOS.}
\begin{tabular}{l|cc|cc||c|c}
\toprule
Task  & \multicolumn{2}{c|}{Circulatory failure} & \multicolumn{2}{c||}{Respiratory failure} &    Kidney func. &       Remaining LOS \\
\midrule
Metric  &      AUPRC ($\uparrow$) & AUROC ($\uparrow$) &  AUPRC ($\uparrow$) & AUROC ($\uparrow$) & MAE ($\downarrow$) &  MAE ($\downarrow$) \\
\midrule
LR                 &          30.5 $\pm$ 0.0 &     87.6 $\pm$ 0.0 &       53.0 $\pm$ 0.0 &      65.4 $\pm$ 0.0 &     N.A &       N.A \\
LGBM               &          \textbf{38.9} $\pm$ 0.3 &     \textbf{91.2} $\pm$ 0.1 &      58.5 $\pm$ 0.1 &     69.3 $\pm$ 0.2 &     \textbf{0.45} $\pm$ 0.00 &   56.9 $\pm$ 0.4 \\
LGBM w. Feat.      &          \textbf{38.8} $\pm$ 0.2 &     \textbf{91.2} $\pm$ 0.1 &      \textbf{60.4} $\pm$ 0.2 &    \textbf{ 70.8} $\pm$ 0.1 &     \textbf{0.45} $\pm$ 0.00 &   57.0 $\pm$ 0.3 \\
\midrule
\midrule
GRU                &         36.8 $\pm$ 0.5 &     90.7 $\pm$ 0.2 &      59.2 $\pm$ 0.3 &     70.1 $\pm$ 0.2 &     0.49 $\pm$ 0.02 &   60.6 $\pm$ 0.9 \\
LSTM               &         32.6 $\pm$ 0.8 &     89.9 $\pm$ 0.1 &      56.9 $\pm$ 0.3 &     68.2 $\pm$ 0.3 &     0.50 $\pm$ 0.01 &  60.7 $\pm$ 1.6 \\
TCN                &         35.8 $\pm$ 0.6 &     90.5 $\pm$ 0.1 &      58.9 $\pm$ 0.3 &     70.0 $\pm$ 0.2 &      0.50 $\pm$ 0.01 &  59.8 $\pm$ 2.8 \\
Transformer        &         35.2 $\pm$ 0.6 &     90.6 $\pm$ 0.2 &      59.4 $\pm$ 0.3 &     70.1 $\pm$ 0.2 &      0.48 $\pm$ 0.02 &       59.5 $\pm$ 2.8 \\
\bottomrule
\end{tabular}
    \label{tab:online-level}
\end{table}

\paragraph{Stay-Level Tasks} When comparing methods on tasks requiring a single prediction after 24h (Table \ref{tab:stay-level}), we observe the superiority of LGBM with hand-extracted features. Transformers outperformed other DL methods but we observe a significant performance gap with the best ML method in B-Accuracy for Patient Phenotyping and AUPRC for ICU Mortality. Concerning GRU and LSTM, their performance is similar to TCN's for ICU Mortality. However, on the Patient Phenotyping task, they do not manage to outperform even logistic regression.

\paragraph{Online Failure Predictions} For the continuous classification tasks, where the maximum sequence length extends from 288 steps to 2016, DL methods do not leverage the additional history information. Indeed, as shown in Table \ref{tab:online-level}, for both Circulatory and Respiratory Failure, LGBM trained only on the current variables outperforms all DL methods. Among these methods, LSTM is the most impacted, as it has noticeably lower scores. Finally, for all continuous tasks, including regression discussed below, the improvement brought by hand-extracted features is not as significant. It suggests that statistical features, when extracted from the entire history, are less informative.

\paragraph{Online Regression Tasks} The final set of tasks we benchmark are regression tasks (Table \ref{tab:online-level}). As for the classification case, LGBM-based methods outperform DL methods, which, among them, have similar performance. In addition, we do not observe any improvement brought by our selection of hand-extracted features. Moreover, the overall performances of the proposed methods are relatively low. While a MAE of $0.45$ ml/kg/h for Kidney Function is only twice smaller than the median urine output rate, a $57$h error in Remaining LOS is more than twice the median length-of-stay. We believe these low scores are due to the nature of the labels' distributions, which are both heavy-tailed as shown in \textsc{Appendix A: Dataset Details}.

\subsection{Behaviour of Deep Learning Approaches for Long Time-Series}
One notable difference between the MIMIC-III benchmark \cite{harutyunyan2019multitask} and our work is the data resolution. The resolution of our data being twelve-time higher leads to 2016 steps (1 week) sequences for online tasks. Thus, we explore if the increase of sequence length explains the decrease in performance of DL methods for continuous tasks.

\begin{figure}[ht!]
    \centering
    \includegraphics[scale=0.41]{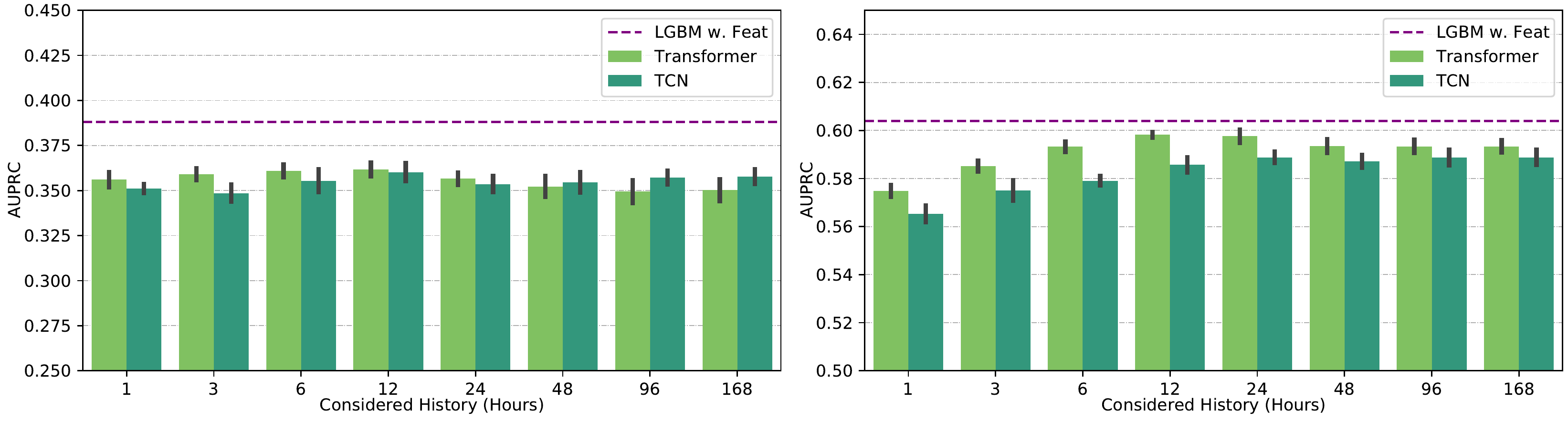}
    \caption{\textit{Impact of history length on online classification performance}. (Left) Comparison in AUPRC for the Circulatory Failure task ; (Right) Comparison in AUPRC for the Respiratory Failure task. Error bars represent the standard deviation over 5 runs with different random initializations.}
    \label{fig:ablation-history}
\end{figure}

\paragraph{History Length} One way to verify if DL methods leverage long-term dependencies in their prediction is to check if a decrease in the considered history impacts performance. We can achieve this for Transformers and TCN architectures, by respectively using local attention or fixing the number of dilated convolutions. In the results (Figure \ref{fig:ablation-history}), we observe that both models do not use the additional information provided by early steps for the Circulatory failure task. It is in line with LGBM's lack of performance improvement when provided with history features on this task. For the Respiratory Failure task, where history features improve LGBM performance, shortening considered history impacts significantly both methods. TCN performance consistently decreases as history diminishes, whereas the Transformer model AUPRC first improves, almost closing the gap with LGBM, before also lowering. Thus, both DL models leverage history in the Respiratory Failure task. However, this also highlights known limitations of Transformers for long sequences \cite{DBLP:journals/corr/abs-1904-10509} when the history exceeds 12h.

\begin{figure}[!ht]
    \centering
    \includegraphics[scale=0.41]{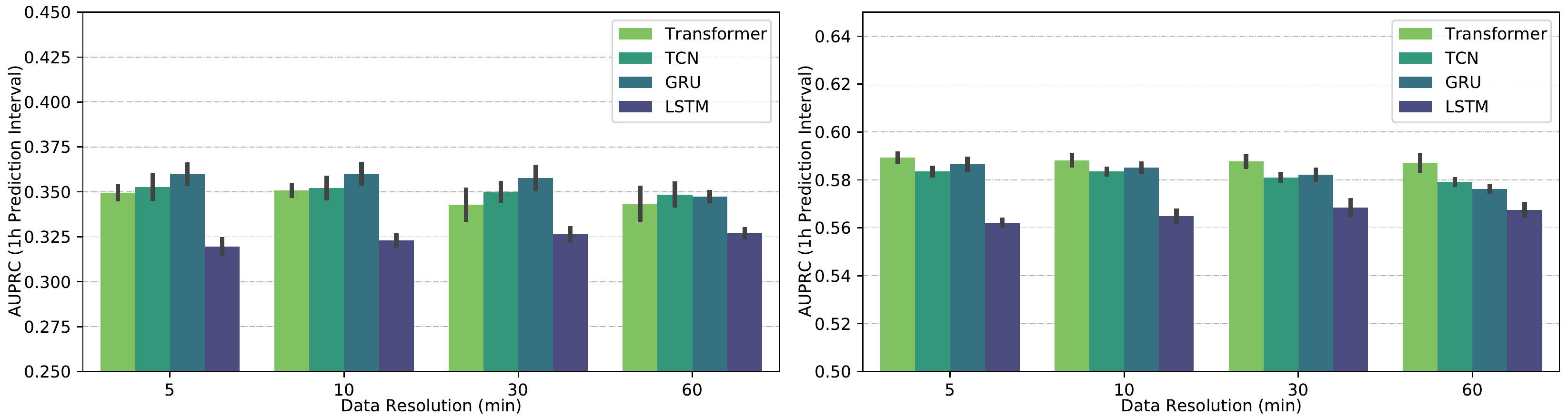}
    \caption{\textit{Impact of data resolution on online classification performance}. (Left) Comparison in AUPRC for the Circulatory Failure task ; (Right) Comparison in AUPRC for the Respiratory Failure task. Error bars represent the standard deviation over 5 runs with different random initializations.}
    \label{fig:ablation-resolution}
\end{figure}

\paragraph{Data Resolution} Another approach to decrease the length of sequences is to reduce the data resolution.  We compare all DL methods with a 1h prediction interval to assess the impact of data resolution on performance. This way, we can gradually lower the data resolution from 5min to 1h while preserving the same prediction time-steps. We report the result of this experiment in Figure \ref{fig:ablation-resolution}. We observe that while TCN and Transformer performance are almost identical, GRU and LSTM are both impacted in opposite ways. GRU is noticeably better than LSTM on both tasks with a 5min grid, but as resolution lowers to 1h, this gap is significantly reduced.

\subsection{On Weighting Cross-entropy by Class Prevalence }\label{ablation:weight}
All the tasks we define show a certain degree of imbalance and the class imbalance problem (CIP) is known to be highly challenging \cite{johnson2019survey}. The most common approach to this problem is the use of class weights in the loss objective. For this ablation, we adopt the original idea from \citep{king2001logistic} by defining weights inversely proportional to each class prevalence. As shown in Table \ref{tab:weighting-impact}, for the multi-class task, it yields a significant improvement on the balanced accuracy. However, such a technique harms all binary classification tasks performances. It is particularly true for the highly imbalanced tasks, Circulatory Failure, and ICU Mortality.

\begin{table}[!ht]
    \centering
    \footnotesize
    \setlength\tabcolsep{1.5pt}
    \caption{\textit{Deltas in metrics of using balanced cross-entropy loss}. (Blue) Improvements over using no weights; (Red) Deterioration over using no weights.   }
\begin{tabular}{l|cc|c|cc|cc}
\toprule
Task & \multicolumn{2}{c|}{ICU Mortality} & Phenotyping & \multicolumn{2}{c|}{Circulatory Failure} & \multicolumn{2}{c}{Respiratory Failure} \\
$\Delta$ Metric & AUPRC ($\uparrow$) & AUROC ($\uparrow$) & B-Accuracy ($\uparrow$) &      AUPRC ($\uparrow$) & AUROC ($\uparrow$) &  AUPRC ($\uparrow$) & AUROC ($\uparrow$) \\
\midrule
LGBM w. Feat. &               \color{red}{-1.3} &                0.0 &                    \color{RoyalBlue}{+4.3} &                    \color{red}{-4.2} &                \color{red}{-2.6} &               \color{RoyalBlue}{0.1} &               0.0 \\
Transformer &               \color{red}{-2.6} &                0.0 &                    \color{RoyalBlue}{+4.0} &                    \color{red}{-0.9} &               0.0 &                \color{red}{-0.1} &               \color{red}{-0.1} \\
\bottomrule
\end{tabular}

    \label{tab:weighting-impact}
\end{table}

 \section{Discussion}\label{disc}

In this paper, we provided an in-depth benchmark on the HiRID dataset and evaluated the behaviour of various machine learning models on diverse clinically relevant tasks developed in collaboration with intensive care clinicians. Our primary contribution is a full and reproducible preprocessing and machine learning pipeline and benchmark tasks on a public intensive care dataset, a necessary prerequisite for reproducible and comparable research in the future. We further evaluate current state-of-the-art machine and deep learning algorithms on these tasks establishing a baseline to compare future methods against. We consider this our second major contribution. 

This work confirms previous results \citep{hyland2020early}, that conventional machine learning models (i.e. boosted ensembles of decision trees) outperform current deep learning approaches on medical time series problems. Based on the experimental results we found that deep learning models do not lead to the same breakthrough performance increases as in other domains (such as NLP~\cite{devlin2018bert} or Computer Vision~\cite{dosovitskiy2020image}). We believe the sparsity of the data and the imbalance of labels in both regression and classification tasks play an important role in this. For classification tasks, building a specific objective for highly imbalanced tasks such as Focal loss \cite{lin2017focal} might be a potential direction of research. For regression, a recent work has shown some promising leads for heavy-tailed regression tasks \cite{yang2021delving}.
Moreover, HiRID introduces a novel high-resolution aspect in ICU data, that needs to be correctly taken into account. Thus, as for other sequence data, one possible explanation could be that when trained with extremely long sequences, models can not use the extracted features in the most effective way~\cite{zaheer2020big}. In the case of Transformers, to force the model to learn and extract useful patterns, various kinds of improvements could be made~\cite{tay2020efficient}. In particular, \textit{learnable patterns} could be incorporated~\cite{roy2021efficient}. 

Our work goes beyond previous ICU time-series benchmarks (e.g. \citep{harutyunyan2019multitask}) by using a more diverse set of tasks and a data set with a higher time resolution. As discussed earlier the set of clinical prediction tasks is diverse regarding the assessed organ systems, prevalence, and task type. An important limitation of our study is that HiRID is currently not the most frequently used and known ICU data set.

This work facilitates the future development of machine learning methods and standardized comparison of their performance on a diverse set of predefined tasks. It could contribute to solving today's problem of machine learning on medical time series not being comparable due to each work's unique datasets, preprocessing, and tasks definition. We hypothesize that methods developed and successfully evaluated on these tasks can also be successfully transferred to other specific medical time series problems.

This work also fills the gap between proposed machine learning approaches and their applications to ICU tasks. As a concrete example, COVID-19 is a big challenge for ICU patient monitoring. Important issues in this context are the uncertainty of the patient's prognosis as well as the prediction of the disease progression. COVID-19 is known to cause respiratory failure \cite{li2020neuroinvasive}, one of the tasks studied in our benchmark, which is also the main cause for ICU admission and death \cite{li2020acute, ruan2020clinical, richardson2020presenting, holter2020systemic}. During the current COVID-19 pandemic, first attempts to construct a Respiratory Failure prediction model were already done such as \cite{bolourani2021machine}, however, their data is available only for a limited audience, limiting
reproducibility.

 \section{Conclusion}
In this paper, we proposed an in-depth benchmark on time series collected from an Intensive Care Unit (ICU). In collaboration with clinicians, we defined several tasks relevant for healthcare covering different critical aspects of ICU patient monitoring. We provide a reproducible end-to-end pipeline to derive both data and labels, and a training setup to evaluate the final performance. 
We hope that this benchmark facilitates the construction and evaluation of machine learning methods for ICU data, and encourages reproducible research in this field.

\bibliographystyle{plain}
\bibliography{references}

\begin{thebibliography}{10}

\bibitem{ams}
AmsterdamUMCdb.
\newblock \url{https://amsterdammedicaldatascience.nl/}.

\bibitem{ARDS_Definition_Task_Force2012-hl}
{ARDS Definition Task Force}, V~Marco Ranieri, Gordon~D Rubenfeld, B~Taylor
  Thompson, Niall~D Ferguson, Ellen Caldwell, Eddy Fan, Luigi Camporota, and
  Arthur~S Slutsky.
\newblock Acute respiratory distress syndrome: the {B}erlin {D}efinition.
\newblock {\em JAMA}, 307(23):2526--2533, June 2012.

\bibitem{bai2018empirical}
Shaojie Bai, J~Zico Kolter, and Vladlen Koltun.
\newblock An empirical evaluation of generic convolutional and recurrent
  networks for sequence modeling.
\newblock {\em arXiv preprint arXiv:1803.01271}, 2018.

\bibitem{bellamy2020evaluating}
David Bellamy, Leo Celi, and Andrew~L Beam.
\newblock Evaluating progress on machine learning for longitudinal electronic
  healthcare data.
\newblock {\em arXiv preprint arXiv:2010.01149}, 2020.

\bibitem{bolourani2021machine}
Siavash Bolourani, Max Brenner, Ping Wang, Thomas McGinn, Jamie~S Hirsch,
  Douglas Barnaby, Theodoros~P Zanos, Northwell {COVID-19}~Research Consortium,
  et~al.
\newblock A machine learning prediction model of respiratory failure within 48
  hours of patient admission for {COVID-19}: model development and validation.
\newblock {\em Journal of medical Internet research}, 23(2):e24246, 2021.

\bibitem{brodersen2010balanced}
Kay~Henning Brodersen, Cheng~Soon Ong, Klaas~Enno Stephan, and Joachim~M
  Buhmann.
\newblock The balanced accuracy and its posterior distribution.
\newblock In {\em 2010 20th international conference on pattern recognition},
  pages 3121--3124. IEEE, 2010.

\bibitem{DBLP:journals/corr/abs-1904-10509}
Rewon Child, Scott Gray, Alec Radford, and Ilya Sutskever.
\newblock Generating long sequences with sparse transformers.
\newblock {\em CoRR}, abs/1904.10509, 2019.

\bibitem{cho2014learning}
Kyunghyun Cho, Bart Van~Merri{\"e}nboer, Caglar Gulcehre, Dzmitry Bahdanau,
  Fethi Bougares, Holger Schwenk, and Yoshua Bengio.
\newblock Learning phrase representations using {RNN} encoder-decoder for
  statistical machine translation.
\newblock {\em arXiv preprint arXiv:1406.1078}, 2014.

\bibitem{citi2012physionet}
Luca Citi and Riccardo Barbieri.
\newblock Physionet 2012 challenge: Predicting mortality of icu patients using
  a cascaded {SVM-GLM} paradigm.
\newblock In {\em 2012 Computing in Cardiology}, pages 257--260. IEEE, 2012.

\bibitem{devlin2018bert}
Jacob Devlin, Ming-Wei Chang, Kenton Lee, and Kristina Toutanova.
\newblock Bert: Pre-training of deep bidirectional transformers for language
  understanding.
\newblock {\em arXiv preprint arXiv:1810.04805}, 2018.

\bibitem{dosovitskiy2020image}
Alexey Dosovitskiy, Lucas Beyer, Alexander Kolesnikov, Dirk Weissenborn,
  Xiaohua Zhai, Thomas Unterthiner, Mostafa Dehghani, Matthias Minderer, Georg
  Heigold, Sylvain Gelly, et~al.
\newblock An image is worth 16x16 words: Transformers for image recognition at
  scale.
\newblock {\em arXiv preprint arXiv:2010.11929}, 2020.

\bibitem{ellis1989determination}
RK~Ellis.
\newblock Determination of po2 from saturation.
\newblock {\em Journal of applied physiology}, 67(2):902--902, 1989.

\bibitem{hirid}
M.~Faltys, M.~Zimmermann, X.~Lyu, M.~Hüser, S.~Hyland, G.~Rätsch, and
  T.~Merz.
\newblock {HiRID}, a high time-resolution icu dataset (version 1.1.1).
\newblock {\em PhysioNet}, 2021.

\bibitem{faouzi2020pyts}
Johann Faouzi and Hicham Janati.
\newblock pyts: A python package for time series classification.
\newblock {\em Journal of Machine Learning Research}, 21(46):1--6, 2020.

\bibitem{goldberger2000physiobank}
Ary~L Goldberger, Luis~AN Amaral, Leon Glass, Jeffrey~M Hausdorff, Plamen~Ch
  Ivanov, Roger~G Mark, Joseph~E Mietus, George~B Moody, Chung-Kang Peng, and
  H~Eugene Stanley.
\newblock {PhysioBank, PhysioToolkit, and PhysioNet}: components of a new
  research resource for complex physiologic signals.
\newblock {\em circulation}, 101(23):e215--e220, 2000.

\bibitem{guecioueurpysf}
Ahmed Guecioueur.
\newblock pysf: Supervised forecasting of sequential data in python, 2018.
\newblock \url{https://github. com/alan-turing-institute/pysf}.

\bibitem{harutyunyan2019multitask}
Hrayr Harutyunyan, Hrant Khachatrian, David~C Kale, Greg Ver~Steeg, and Aram
  Galstyan.
\newblock Multitask learning and benchmarking with clinical time series data.
\newblock {\em Scientific data}, 6(1):1--18, 2019.

\bibitem{hochreiter1997long}
Sepp Hochreiter and J{\"u}rgen Schmidhuber.
\newblock Long short-term memory.
\newblock {\em Neural computation}, 9(8):1735--1780, 1997.

\bibitem{holter2020systemic}
Jan~C Holter, Soeren~E Pischke, Eline de~Boer, Andreas Lind, Synne Jenum,
  Aleksander~R Holten, Kristian Tonby, Andreas Barratt-Due, Marina Sokolova,
  Camilla Schjalm, et~al.
\newblock Systemic complement activation is associated with respiratory failure
  in {COVID-19} hospitalized patients.
\newblock {\em Proceedings of the National Academy of Sciences},
  117(40):25018--25025, 2020.

\bibitem{horn2020set}
Max Horn, Michael Moor, Christian Bock, Bastian Rieck, and Karsten Borgwardt.
\newblock Set functions for time series.
\newblock In {\em International Conference on Machine Learning}, pages
  4353--4363. PMLR, 2020.

\bibitem{hyland2020early}
Stephanie~L Hyland, Martin Faltys, Matthias H{\"u}ser, Xinrui Lyu, Thomas
  Gumbsch, Crist{\'o}bal Esteban, Christian Bock, Max Horn, Michael Moor,
  Bastian Rieck, et~al.
\newblock Early prediction of circulatory failure in the intensive care unit
  using machine learning.
\newblock {\em Nature medicine}, 26(3):364--373, 2020.

\bibitem{jarrettclairvoyance}
Daniel Jarrett, Jinsung Yoon, Ioana Bica, Zhaozhi Qian, Ari Ercole, and Mihaela
  van~der Schaar.
\newblock Clairvoyance: A pipeline toolkit for medical time series.
\newblock In {\em International Conference on Learning Representations}, 2020.

\bibitem{MIMIC-IV}
Alistair Johnson, Lucas Bulgarelli, Tom Pollard, Steven Horng, Leo~Anthony
  Celi, and Roger Mark.
\newblock {MIMIC-IV}, 2020.

\bibitem{johnson2017reproducibility}
Alistair~EW Johnson, Tom~J Pollard, and Roger~G Mark.
\newblock Reproducibility in critical care: a mortality prediction case study.
\newblock In {\em Machine Learning for Healthcare Conference}, pages 361--376.
  PMLR, 2017.

\bibitem{johnson2016mimic}
Alistair~EW Johnson, Tom~J Pollard, Lu~Shen, H~Lehman Li-Wei, Mengling Feng,
  Mohammad Ghassemi, Benjamin Moody, Peter Szolovits, Leo~Anthony Celi, and
  Roger~G Mark.
\newblock {MIMIC-III}, a freely accessible critical care database.
\newblock {\em Scientific data}, 3(1):1--9, 2016.

\bibitem{johnson2019survey}
Justin~M Johnson and Taghi~M Khoshgoftaar.
\newblock Survey on deep learning with class imbalance.
\newblock {\em Journal of Big Data}, 6(1):1--54, 2019.

\bibitem{ke2017lightgbm}
Guolin Ke, Qi~Meng, Thomas Finley, Taifeng Wang, Wei Chen, Weidong Ma, Qiwei
  Ye, and Tie-Yan Liu.
\newblock {LightGBM}: A highly efficient gradient boosting decision tree.
\newblock {\em Advances in neural information processing systems},
  30:3146--3154, 2017.

\bibitem{king2001logistic}
Gary King and Langche Zeng.
\newblock Logistic regression in rare events data.
\newblock {\em Political analysis}, 9(2):137--163, 2001.

\bibitem{kingma2014adam}
Diederik~P Kingma and Jimmy Ba.
\newblock Adam: A method for stochastic optimization.
\newblock {\em arXiv preprint arXiv:1412.6980}, 2014.

\bibitem{Knaus1985-iw}
W~A Knaus, E~A Draper, D~P Wagner, and J~E Zimmerman.
\newblock {APACHE} {II}: a severity of disease classification system.
\newblock {\em Crit. Care Med.}, 13(10):818--829, October 1985.

\bibitem{li2020acute}
Xu~Li and Xiaochun Ma.
\newblock Acute respiratory failure in {COVID-19}: is it “typical” ards?
\newblock {\em Critical Care}, 24:1--5, 2020.

\bibitem{li2020neuroinvasive}
Yan-Chao Li, Wan-Zhu Bai, and Tsutomu Hashikawa.
\newblock The neuroinvasive potential of {SARS-CoV2} may play a role in the
  respiratory failure of {COVID-19} patients.
\newblock {\em Journal of medical virology}, 92(6):552--555, 2020.

\bibitem{lin2017focal}
Tsung-Yi Lin, Priya Goyal, Ross Girshick, Kaiming He, and Piotr Doll{\'a}r.
\newblock Focal loss for dense object detection.
\newblock In {\em Proceedings of the IEEE international conference on computer
  vision}, pages 2980--2988, 2017.

\bibitem{paszke2019pytorch}
Adam Paszke, Sam Gross, Francisco Massa, Adam Lerer, James Bradbury, Gregory
  Chanan, Trevor Killeen, Zeming Lin, Natalia Gimelshein, Luca Antiga, et~al.
\newblock {PyTorch}: An imperative style, high-performance deep learning
  library.
\newblock {\em arXiv preprint arXiv:1912.01703}, 2019.

\bibitem{scikit-learn}
F.~Pedregosa, G.~Varoquaux, A.~Gramfort, V.~Michel, B.~Thirion, O.~Grisel,
  M.~Blondel, P.~Prettenhofer, R.~Weiss, V.~Dubourg, J.~Vanderplas, A.~Passos,
  D.~Cournapeau, M.~Brucher, M.~Perrot, and E.~Duchesnay.
\newblock Scikit-learn: Machine learning in {P}ython.
\newblock {\em Journal of Machine Learning Research}, 12:2825--2830, 2011.

\bibitem{pollard2018eicu}
Tom~J Pollard, Alistair~EW Johnson, Jesse~D Raffa, Leo~A Celi, Roger~G Mark,
  and Omar Badawi.
\newblock The {eICU} collaborative research database, a freely available
  multi-center database for critical care research.
\newblock {\em Scientific data}, 5(1):1--13, 2018.

\bibitem{reyna2019early}
Matthew~A Reyna, Chris Josef, Salman Seyedi, Russell Jeter, Supreeth~P
  Shashikumar, M~Brandon Westover, Ashish Sharma, Shamim Nemati, and Gari~D
  Clifford.
\newblock Early prediction of sepsis from clinical data: the
  physionet/computing in cardiology challenge 2019.
\newblock In {\em 2019 Computing in Cardiology (CinC)}, pages Page--1. IEEE,
  2019.

\bibitem{richardson2020presenting}
Safiya Richardson, Jamie~S Hirsch, Mangala Narasimhan, James~M Crawford, Thomas
  McGinn, Karina~W Davidson, Douglas~P Barnaby, Lance~B Becker, John~D Chelico,
  Stuart~L Cohen, et~al.
\newblock Presenting characteristics, comorbidities, and outcomes among 5700
  patients hospitalized with {COVID-19} in the {N}ew {Y}ork {C}ity area.
\newblock {\em Jama}, 323(20):2052--2059, 2020.

\bibitem{roy2021efficient}
Aurko Roy, Mohammad Saffar, Ashish Vaswani, and David Grangier.
\newblock Efficient content-based sparse attention with routing transformers.
\newblock {\em Transactions of the Association for Computational Linguistics},
  9:53--68, 2021.

\bibitem{ruan2020clinical}
Qiurong Ruan, Kun Yang, Wenxia Wang, Lingyu Jiang, and Jianxin Song.
\newblock Clinical predictors of mortality due to {COVID-19} based on an
  analysis of data of 150 patients from wuhan, china.
\newblock {\em Intensive care medicine}, 46(5):846--848, 2020.

\bibitem{sadeghi2018early}
Reza Sadeghi, Tanvi Banerjee, and William Romine.
\newblock Early hospital mortality prediction using vital signals.
\newblock {\em Smart Health}, 9:265--274, 2018.

\bibitem{tay2020efficient}
Yi~Tay, Mostafa Dehghani, Dara Bahri, and Donald Metzler.
\newblock Efficient transformers: A survey.
\newblock {\em arXiv preprint arXiv:2009.06732}, 2020.

\bibitem{tomavsev2019clinically}
Nenad Toma{\v{s}}ev, Xavier Glorot, Jack~W Rae, Michal Zielinski, Harry Askham,
  Andre Saraiva, Anne Mottram, Clemens Meyer, Suman Ravuri, Ivan Protsyuk,
  et~al.
\newblock A clinically applicable approach to continuous prediction of future
  acute kidney injury.
\newblock {\em Nature}, 572(7767):116--119, 2019.

\bibitem{vaswani2017attention}
Ashish Vaswani, Noam Shazeer, Niki Parmar, Jakob Uszkoreit, Llion Jones,
  Aidan~N Gomez, Lukasz Kaiser, and Illia Polosukhin.
\newblock Attention is all you need.
\newblock {\em arXiv preprint arXiv:1706.03762}, 2017.

\bibitem{yang2021delving}
Yuzhe Yang, Kaiwen Zha, Ying-Cong Chen, Hao Wang, and Dina Katabi.
\newblock Delving into deep imbalanced regression.
\newblock {\em arXiv preprint arXiv:2102.09554}, 2021.

\bibitem{ncle}
Hugo Yèche, Gideon Dredsner, Francesco Locatello, Matthias Hüser, and Gunnar
  Rätsch.
\newblock Neighborhood contrastive learning applied to online patient
  monitoring.
\newblock In {\em International Conference on Machine Learning}. PMLR, 2021.

\bibitem{zaheer2020big}
Manzil Zaheer, Guru Guruganesh, Avinava Dubey, Joshua Ainslie, Chris Alberti,
  Santiago Ontanon, Philip Pham, Anirudh Ravula, Qifan Wang, Li~Yang, et~al.
\newblock {Big Bird}: Transformers for longer sequences.
\newblock {\em arXiv preprint arXiv:2007.14062}, 2020.

\bibitem{Zimmerman2006-of}
Jack~E Zimmerman, Andrew~A Kramer, Douglas~S McNair, and Fern~M Malila.
\newblock Acute physiology and chronic health evaluation ({APACHE}) {IV}:
  hospital mortality assessment for today's critically ill patients.
\newblock {\em Crit. Care Med.}, 34(5):1297--1310, May 2006.

\end{thebibliography}

\appendix

\section{Appendix A: HiRID Dataset Details}

\section*{Dataset description}

The HiRID ICU data is provided by the University Hospital Bern, Bern, Switzerland.
The dataset consists of 33,905 patients whose length of stays in the ICU range from 0 to 28 days. A more detailed summary of the HiRID cohort statistics can be found in Table~\ref{tab:cohort_stats}.

For all patients, a variety of clinical measurements are collected. It represents a set of 710 variables that can be categorized into the following types:
\begin{itemize}
    \item Demographics (e.g.: Sex, Age, Height)
    \item Bedside vital signs (e.g.: Heart rate)
    \item Settings of medical devices (e.g.: Mechanical ventilation)
    \item Manual observations (e.g.: Urine output)
    \item Lab measurements (e.g.: Lactate)
    \item Treatments (e.g.: Vasopressor agents)
\end{itemize}

One notable difference between these types of measurements is the resolution at which they are provided. Bedside vital signs are provided at regular intervals of 2 min, whereas lab measurements are only available every couple of hours at best. The frequency of recording discrepancy existing between different variables yields unique challenges for machine learning models. Further details about the available clinical measurements can be found in the official documentation of HiRID\footnote{URL for the official HiRID documentation \url{https://hirid.intensivecare.ai/}} as well as in our software repository\footnote{ Table containing all information for each variable: \url{https://github.com/ratschlab/HIRID-ICU-Benchmark/blob/master/preprocessing/resources/varref.tsv}}

\begin{table}[!ht]
    \centering
    \caption{Cohort statistics of the public HiRID dataset v1.1.1, as released on Physionet, 
    which was used for the HiRID-ICU benchmark}\label{tab:cohort_stats}
    {\fontfamily{cmss}\selectfont\scriptsize
    \begin{tabular}{l l l r l}
    \toprule
         \multirow{2}{*}{\textbf{Sex}} & \multicolumn{2}{l}{Female} & 12,138 & patients \\ \cmidrule{2-5}
         & \multicolumn{2}{l}{Male} & 21,767 & patients \\ \midrule
         \multirow{8}{*}{\textbf{Age group}} & \multicolumn{2}{l}{[20,30)} & 1,042 & patients \\ \cmidrule{2-5}
         & \multicolumn{2}{l}{[30,40)} & 1,278 & patients \\ \cmidrule{2-5}
         & \multicolumn{2}{l}{[40,50)} & 2,649 & patients \\ \cmidrule{2-5}
         & \multicolumn{2}{l}{[50,60)} & 5,194 & patients \\ \cmidrule{2-5}
         & \multicolumn{2}{l}{[60,70)} & 8,241 & patients \\ \cmidrule{2-5}
         & \multicolumn{2}{l}{[70,80)} & 9,445 & patients \\ \cmidrule{2-5}
         & \multicolumn{2}{l}{[80,90)} & 5,534 & patients \\ \cmidrule{2-5}
         & \multicolumn{2}{l}{[90,100)} & 522 & patients \\ \midrule
         \multirow{3}{*}{\textbf{Discharge status}} & \multicolumn{2}{l}{Alive} & 31,604 & patients \\ \cmidrule{2-5}
         & \multicolumn{2}{l}{Dead} & 2,062 & patients \\ \cmidrule{2-5}
         & \multicolumn{2}{l}{Unknown} & 239 & patients \\ \midrule
         \multirow{15}{*}{\makecell{\textbf{APACHE group} \\ \textbf{(Patient phenotype)}}} & \multirow{2}{*}{Cardiovascular} & Surgical & 8,125& patients \\ \cmidrule{3-5}
         & & Non-surgical & 4,356 & patients \\ \cmidrule{2-5}
         & \multirow{2}{*}{Neurologic} & Surgical & 3,938 & patients \\ \cmidrule{3-5}
         & & Non-surgical & 6,014 & patients \\ \cmidrule{2-5}
         & \multirow{2}{*}{Gastrointestinal} & Surgical & 1,768 & patients \\ \cmidrule{3-5}
         & & Non-surgical & 1,918 & patients \\ \cmidrule{2-5}
         & \multirow{2}{*}{Respiratory} & Surgical & 631 & patients \\ \cmidrule{3-5}
         & & Non-surgical & 2,399 & patients \\ \cmidrule{2-5}
         & \multirow{2}{*}{Trauma} & Surgical & 239 & patients \\ \cmidrule{3-5}
         & & Non-surgical & 1,522 & patients \\ \cmidrule{2-5}
         & \multicolumn{2}{l}{Other medical diseases} & 1,428 & patients \\ \cmidrule{2-5}
         & \multicolumn{2}{l}{Other surgical} & 568 & patients \\ \cmidrule{2-5}
         & \multicolumn{2}{l}{Metabolic/Endocrinology} & 630 & patients \\ \cmidrule{2-5}
         & \multicolumn{2}{l}{Hematologic} & 99 & patients \\ \cmidrule{2-5}
         & \multicolumn{2}{l}{Renal surgical} & 81 & patients \\ \cmidrule{2-5}
         & \multicolumn{2}{l}{Unknown} & 189 & patients \\ \midrule
\multirow{2}{*}{\textbf{Length of stay }} & \multicolumn{2}{l}{Median} & 0.95 & days \\ \cmidrule{2-5}
         & \multicolumn{2}{l}{Range} & (0,28] & days \\ \midrule
    \end{tabular}
    }
\end{table}

\section*{Data preprocessing}\label{appsec:data_preprocessing}

\paragraph{Artifact removal} The HiRID data when directly exported from the data management system contains various types of artifacts. The most common one concerns measurement values that are out of the normal range defined by clinicians. Another artifact that we observed is that the time of the first measurement record of cumulative variables is at noon by default, which could be much earlier than the admission time of the patient. This, if not taken into account correctly, will affect the rate calculation for those cumulative variables.
To solve these timestamp issues, we clip the time of the first record of 
the cumulative variables to the ICU admission time of the patient.

\paragraph{Variable merging} There are often various variables in the ICU EHRs that have the same or similar clinical meaning. As an example, there exist three variables for temperature, namely core body temperature, auxiliary temperature, and rectal temperature. 
Usually, only one or few variables from one medical concept are measured for a patient, therefore, merging variables with the same clinical concept into one meta-variable reduces the sparsity of the feature matrices.
Another advantage is that machine learning models trained on medical concept features are more transferable to other hospitals than those trained on the original clinical variables because hospitals usually have their own variable coding scheme.  We hence incorporated variable-merging in our first pre-processing step. 
A challenge that arises from merging pharmaceutical variables is the dosage normalization across drugs with similar active ingredients.
Therefore, instead of using the drug dosage as features, which likely contain invalid information, we use ``presence/absence'' binary indicators of drugs as features for the machine learning models.

\paragraph{Definition of pharmaceutical acting periods} When a patient has been administrated a drug at a time point, only this time point contains a measurement. However, each drug is active for a specific duration after this administration time. For this reason, we ensure to propagate the measurement for a duration corresponding to the drug acting period. We set the value back to zero after this period ensuring compatibility with forward filling imputation. 

\paragraph{Conversion of cumulative values to rates} There are a few fluid output variables whose measurement values are recorded as cumulative every 12 hours starting from noon on the ICU admission day. Since the cumulative signals are periodic, we converted them to hourly fluid rates which are more interpretable and amenable as machine learning features and for the endpoint
definition.

\section*{Endpoint definition and statistics}

\subsection*{Endpoint extraction algorithms}

Our benchmark suite contains a range of clinically relevant tasks chosen to range over a spectrum of different properties of tasks, e.g. task type from the machine learning point-of-view (regression, binary classification, multi-class classification), point of the ICU stay at which the prediction is made (fixed time-point vs. predictions made throughout the stay), as well as the degree of class imbalance (highly imbalanced tasks vs. more balanced tasks). While in the main paper we describe the main idea of each task, the full definition and implementation details of the extraction algorithm are described below.

\paragraph{Dependencies on previous pipeline stages.}Static information about the patient is extracted from the \texttt{static.parquet} file, in which the mortality status and the APACHE-II/IV codes for the admission are stored. This file is derived from the \texttt{general\_table.parquet} and \texttt{observations} tables, where APACHE II/IV codes are stored, during the data preprocessing pipeline stage. Respiratory/kidney/circulatory failure labels are generated from the merged stage of the data, via an intermediate imputation step, in which measurements are forward filled to statistics about the number of real measurements in time grid intervals were pre-computed. These simplified the implementation of the endpoints, whose extraction algorithms use a combination of imputed values and real measurement detection strategies.

\paragraph{Label masking on presence of vital sign monitoring. } All tasks described below were additionally masked with a condition on a current connection to vital sign monitors. This implied setting a valid label (according to the below task definitions) to invalid (\texttt{NaN}) if the patient had no heart rate measurement (\texttt{vm1}) in the 15 minutes surrounding the time grid point. Since the expected frequency of heart rate measurements is 1 sample / 2 minutes, it is likely the patient is disconnected when this condition is satisfied, and no reliable predictions on their state can be made.

\paragraph{Patient phenotyping.}
To derive the APACHE diagnostic group of the patient, the two fields \texttt{APACHE-II group} and \texttt{APACHE-IV group} of the static HiRID data are used. To map these two fields to a standardized encoding, the conversion Table \ref{tab:appendix-apache-map-table} is used. If the patient had a valid \texttt{APACHE-II group} entry and its code was mapped in the table, the target \texttt{APACHE group} code was used. Otherwise, we tried to fall back to the \texttt{APACHE-IV group} entry, where again if it was present and mapped, the corresponding target \texttt{APACHE group} code was used. If still no code could be extracted, the prediction task was defined as invalid for the patient. The label of the time-point 24h after ICU admission was set to the class category of the \texttt{APACHE group}, defining a multi-class classification problem to be solved once in the stay. If the admission was shorter than 24h, no label for the patient was assigned.

\begin{table}[ht!]
\caption{Mapping between APACHE II and IV group codes to one 
         unique APACHE group diagnostic group encoding, which is used for 
         the patient phenotyping task.}
\label{tab:appendix-apache-map-table}
{\fontfamily{cmss}\selectfont\small
\begin{tabular}{lllll}
\toprule
\makecell{\textbf{APACHE II} \\ \textbf{Code}} & \makecell{\textbf{APACHE IV} \\ \textbf{Code}} & \textbf{Original group name}                                           & \textbf{Target group} & \textbf{Target group name}                              \\
\midrule
98                         & 190                        & Cardiovascular                                          & 1                     & {\color[HTML]{333333} Cardiovascular}                   \\
99                         & 191                        & Respiratory                                             & 2                     & Respiratory                                             \\
100                        & 192                        & Gastrointestinal                                        & 3                     & Gastrointestinal                                        \\
101                        & 193                        & Neurologic                                              & 4                     & Neurologic                                              \\
                           & 197                        & Urogenital                                              & 6                     & Other medical diseases                                  \\
102                        &                            & Sepsis                                                  & 6                     & Other medical diseases                                  \\
106                        & 198                        & Other                                                   & 6                     & Other medical diseases                                  \\
                           & 206                        & Intoxication                                            & 6                     & Other medical diseases                                  \\
103                        & 194                        & Trauma not surgical                                     & 7                     & Trauma not surgical                                     \\
104                        & 195                        & Metabolic/Endocrinology                                 & 8                     & Metabolic/Endocrinology                                 \\
105                        & 196                        & Hematologic                                             & 9                     & Hematologic                                             \\
107                        & 199                        & {\color[HTML]{333333} (Cardio)vascular surgical} & 11                    & {\color[HTML]{333333} (Cardio)vascular surgical} \\
108                        & 201                        & Respiratory surgical                                    & 12                    & Respiratory surgical                                    \\
109                        & 200                        & Gastrointestinal surgical                               & 13                    & Gastrointestinal surgical                               \\
110                        & 202                        & Neurologic surgical                                     & 14                    & Neurologic surgical                                     \\
111                        & 203                        & Trauma surgical                                         & 15                    & Trauma surgical                                         \\
112                        & 204                        & Renal surgical                                          & 16                    & Renal surgical                                          \\
113                        &                            & Gynacology surgical                                     & 17                    & Other surgical                                          \\
114                        &                            & Orthopedics surgical                                    & 17                    & Other surgical                                          \\
                           & 205                        & Other surgical                                          & 17                    & Other surgical                                          \\
                           &                            & Unknown                                                 & 18                    & Unknown     \\                                           
\bottomrule
\end{tabular}
}
\end{table}
\FloatBarrier

\paragraph{Mortality.}First, the ICU mortality label was extracted from the general data table of the HiRID database. 
The label of the time-point 24 hours after ICU admission was set to 1 (positive) if the patient died at the end of the stay according
to this field, and 0 (negative) otherwise, defining a binary classification problem to be solved once per stay. 
If the admission was shorter than 24 hours, no label was assigned to the patient. 

\paragraph{Remaining length of stay.}The continuous regression label of a time-point in the ICU stay was defined as the number of hours that remain until the end-of-the-stay, 
i.e. the first label of the ICU stay is equal to the ICU stay length (in hours) and the last label just before dispatch is defined as 0. To determine the beginning and end of the ICU stay, the first and last observed heart rate measurements (\texttt{vm1}) were used.

\paragraph{Kidney function.}As a basis for extracting the kidney function label, the urine output rate (\texttt{vm24}) variable and the weight variable (\texttt{vm131}) were used. The valid labels were anchored to real measurements, i.e. updates, of the urine output rate. Only time-points exactly 2h before an update of the urine output rate were assigned a prediction label. The overall urine output in the 2h after this time point was found by computing the integral of the urine output rate variable (\texttt{vm24}) over this 2h period. The overall urine output in the time interval was then normalized to the weight of the patient (unit \texttt{kg}) at the time of the prediction, and
then standardized to a rate/hour. Hence, the unit of the regression output is 
\texttt{ml/kg/h}.

\paragraph{Circulatory failure.}As a basis for computing the circulatory failure status of a patient at every time-point of the ICU stay, the following variables were used: Mean arterial pressure (\texttt{vm5}), arterial lactate (\texttt{vm136}) and the vasopressive agents Milrinone (\texttt{pm42}), Dobutamine (\texttt{pm41}), Levosimendan (\texttt{pm43}), Theophylline (\texttt{pm44}), Norepinephrine (\texttt{pm39}), Epinephrine (\texttt{pm40}) and Vasopressin (\texttt{pm45}). For defining the label at time $t$ a centralized window of size 2h was anchored at $t$, and the lactate/MAP conditions in this window were separately analyzed. The time-point was assigned a (tentative) positive label if, for at least 2/3 of the time-points inside the 2h, the lactate condition was satisfied, and for at least 2/3 of the time-points inside the 2h, the MAP condition was satisfied. 
The time-points at which they have to be satisfied do not necessarily have to coincide. The conditions defining the circulatory failure state are listed below:

\begin{itemize}
\item \textbf{Lactate condition}. Elevated arterial lactate ($>2$ mmol/l)
\item \textbf{MAP/vasopressor condition}. Low MAP ($<65$ mmHg ) or any dose 
of any of the considered vasopressors (\texttt{pm39, pm40, pm41, pm42, pm43, pm44, pm45})
given to the patient, which could potentially mask a low MAP.
\end{itemize}

Using these endpoint annotations of ``circulatory failure'' and ``no circulatory failure'', a label at time point $t$ is either (1) \textit{invalid} (no prediction made), if the patient was in circulatory failure at $t$, (2) \textit{positive} if the patient was not circulatory failure at $t$, but there was a new onset of circulatory failure in the next 12h, and (3) \textit{negative} if the patient was not in circulatory failure at $t$, and there was no onset of circulatory failure in the next 12h.

\paragraph{Respiratory failure.}Estimating the respiratory failure status of a patient at a given time-point involves estimating (1) their instantaneous FiO$_2$ value, (2) their instantaneous PaO$_2$ value and computing the P/F ratio as P/F = PaO$_2$/FiO$_2$, and then labeling the time series according to the threshold P/F = 300 mmHg, to define (mild) failure and stability periods.

We distinguished between the following cases to estimate the FiO$_2$ value at a time-point $t$. 

\begin{itemize}
    \item FiO$_2$ from ventilator: If there was a FiO$_2$ measurement (\texttt{vm58}) in the last 30 min and the patient was on the ventilator or the ventilation mode NIV (\texttt{vm60}) was active, then the last FiO$_2$ measurement (\texttt{vm58}) was directly used as the estimate.
    \item FiO$_2$ from supplementary oxygen (oxygen mask): If there was a real measurement in supplementary oxygen (\texttt{vm23}) in the last 12h, then it was forward filled and used to estimate FiO$_2$ via the conversion Table     \ref{tab:appendix-suppox-conv-table}.
    \item FiO$_2$ from ambient air: An ambient air FiO$_2$ assumption was made,
    and FiO$_2$ was estimated as 21 \%.
\end{itemize}

\begin{table}[!ht]
    \caption{ Supplemental Oxygen to FiO$_2$ conversion table used for determining the continuous FiO$_2$ estimate.}
    \label{tab:appendix-suppox-conv-table}
    \begin{center}
        {\fontfamily{cmss}\selectfont\small
            \begin{tabular}{ll}
                \toprule
                \textbf{Supp. oxygen [l]} & \textbf{FiO$_2$ [\%]} \\
                \midrule
                1  & 26 \\
                2 & 34 \\
                3 & 39 \\
                4 & 45 \\
                5 & 49 \\
                6 & 54 \\
                7 & 57 \\
                8 & 58 \\
                \bottomrule
            \end{tabular}
            \begin{tabular}{ll}
                \toprule
                \textbf{Supp. oxygen [l]} & \textbf{FiO$_2$ [\%]} \\
                \midrule
                9 & 63 \\
                10 & 66 \\
                11 & 67 \\
                12 & 69 \\
                13 & 70 \\
                14 & 73 \\
                15 & 75 \\
                $>$15 & 75 \\
                \bottomrule
            \end{tabular}}
    \end{center}
\end{table}
\FloatBarrier

We distinguished between the following cases tor estimating the PaO$_2$ value at a time-point. As source variables peripheral oxygen saturation (\texttt{vm20}) and PaO$_2$ from ABGA (\texttt{vm140}) were used. Hereby the SpO2 variable was pre-smoothed with a percentile (75 \% percentile) moving window filter of size 30 min to remove spuriously low outlier measurements.

\begin{itemize}
    \item Real PaO$_2$ measurement: If there was a real PaO$_2$ (\texttt{vm140}) from an arterial blood gas analysis available in the last 30 minutes, then the estimate was defined directly by the measurement.
    \item Ellis estimate from SpO$_2$: Otherwise the last measurement of peripheral oxygen saturation measured by pulse oximetry (\texttt{vm20}) was used to compute an estimate of PaO$_2$ according to Ellis et al. \cite{ellis1989determination}. If there is no real SpO$_2$ measurement in the last 24h, a default value of 98 \% was assumed for the estimation.
\end{itemize}

First, the resulting PaO$_2$ estimate was smoothed with a Nadaraya-Watson kernel smoother with a bandwidth of 20 min. Then, each so-derived PaO$_2$ estimate was weighted by a squared exponential kernel and converted to the closest plausible PaO$_2$ measurement. The scale of the kernel function was chosen such that  a1h distance with the measurement time resulted in a weight of $\frac{1}{3}$.

The time series was labeled for respiratory failure by using forward-facing windows anchored at time-point $t$. A patient is considered as being in respiratory failure at $t$ if, for at least 2/3 time-points of the next 2h, either of the following conditions hold:

\begin{itemize}
    \item The estimated P/F ratio is $<300$ mmHg and the patient is not on mechanical ventilation \textbf{OR} \item The estimated P/F ratio is $<300$ mmHg and the patient is on mechanical ventilation, but the PEEP value is not densely measured (no real measurement in 30 min) \textbf{OR} 
\item The estimated P/F ratio is $<300$ mmHg and the patient is on mechanical ventilation, and the PEEP is densely measured and the PEEP is $>4$ mmHg.
\end{itemize}

After the definition of event periods, their edges were corrected according to estimates of the P/F ratio at the given time points. If before the event P/F$<300$ mmHg, the event is further extended in the past.  Likewise, if the P/F$>300$ mmHg condition holds after the event, it is extended into the future.

Finally, small events shorter than 4h, preceded and succeeded by two stability periods, of which one is longer than 4h are deleted. This is because these likely represent intermittent/spurious instances of respiratory failure. Conversely, short stability periods shorter than 4h, preceded and followed by periods of respiratory failure, of which one is longer than 4h, are defined as respiratory failure. Likely, during these periods, the patient does not recover from the failure in a clinical sense.

\subsection*{Statistics on annotated failure events and labels}

Below we display statistics on the data resulting from the endpoint stage, and from the label stage, which was used directly as inputs for the machine learning models.

\subsubsection*{Respiratory failure}

\begin{table}[ht!]
\caption{Summary statistics about annotated respiratory failure (oxygenation failure with a P/F ratio <300 mmHg) events in the pre-processed data-set. The
label prevalence refers to the observed probability of developing respiratory failure in the next 12h, given that the patient is currently stable.}
\footnotesize
\begin{center}
{\fontfamily{cmss}\selectfont\small
\begin{tabular}{lll}
\toprule
\textbf{Respiratory failure} & & \\
\midrule
Patients with events & 83.08 \%  & 28,157 admissions \\
Per-time prevalence of resp. failure & 61.83 \% & \\ 
Median \# events per patient (if $\geq$ 1 event) [IQR] & 1 [1-2] & \\
Median event duration [IQR] & 9.5 [4.1-20.6] hours & \\
Median time to first event (if $\geq$ 1 event) [IQR] & 1.58 [0-11] hours  & \\
Median gap length between two events [IQR] & 4.4 [2.1-11] hours & \\
\midrule
Overall label prevalence (Stable->Failure 12h) & 8.6 \% & \\
Median label prevalence p/patient (if $\geq$ 1 event) [IQR] & 0 [0-60.1] \% \\
\bottomrule
\end{tabular}}
\end{center}
\label{tab:appendix-resp-failure-statistics}
\end{table}

\FloatBarrier

\subsubsection*{Circulatory failure}

\begin{table}[ht!]
\caption{Summary statistics about annotated circulatory failure (elevated lactate and abnormally low MAP or vasopressive agents) events in the pre-processed data-set. The label prevalence refers to the observed probability of developing circulatory failure in the next 12h, given that the patient is currently stable.}
\footnotesize
\begin{center}
{\fontfamily{cmss}\selectfont\small
\begin{tabular}{lll}
\toprule
\textbf{Circulatory failure} & & \\
\midrule
Patients with events & 25.58 \%  & 8,669 admissions \\
Per-time prevalence of circ. failure & 6.70 \% & \\ 
Median \# events per patient (if $\geq$ 1 event) [IQR] & 1 [1-2] & \\
Median event duration [IQR] & 2.9 [1.1-7.6] hours & \\
Median time to first event (if $\geq$ 1 event) [IQR] & 1.9 [0.5-7.3] hours  & \\
Median gap length between two events [IQR] & 3.3 [1.3-9.6] hours & \\
\midrule
Overall label prevalence (Stable->Failure 12h) & 1.4 \% & \\
Median label prevalence p/patient (if $\geq$ 1 event) [IQR] & 3.8 [0-19.1] \%  \\
\bottomrule
\end{tabular}}
\end{center}
\label{tab:appendix-resp-failure-statistics}
\end{table}

\FloatBarrier

\subsubsection*{Mortality prediction / Patient phenotyping}

\begin{table}[ht!]
\caption{Summary statistics about the mortality prediction and
patient phenotyping tasks defined at 24h after ICU admission. The last
column shows the number of admissions assigned this label.}
\footnotesize
\begin{center}
{\fontfamily{cmss}\selectfont\small
\begin{tabular}{lll}
\toprule
Percentage of ICU stays with length >24h & 44.1 \% & 14,962 admissions \\
\midrule
\textbf{Mortality prediction} & & \\
\midrule
Overall label prevalence (Mortality at 24 hours) & 8.38 \% & \\
\midrule
\textbf{Patient phenotyping} & & \\
\midrule
Prevalence 'Neurologic' (4) & 23.6 \% & 3,511 admissions \\
Prevalence '(Cardio)vascular' (1) & 15.9 \% & 2,367 admissions \\
Prevalence '(Cardio)vascular surgical (11) & 12.4 \% & 1,848 admissions \\ 
Prevalence 'Respiratory' (2) & 11.0 \% & 1,635 admissions \\
Prevalence 'Neurologic surgical' (14) & 7.4 \% & 1,106 admissions \\
Prevalence 'Trauma not surgical' (7) & 6.7 \% & 990 admissions \\
Prevalence 'Gastrointestinal' (3) & 6.3 \% & 930 admissions \\ 
Prevalence 'Other medical disease' (6) & 5.2 \% & 778 admissions \\ 
Prevalence 'Gastrointestinal surgical' (13) & 4.8 \% & 713 admissions \\
Prevalence 'Metabolic/Endocrinology' (8) & 2.3 \% & 344 admissions \\ 
Prevalence 'Respiratory surgical' (12) & 1.7 \% & 246 admissions \\
Prevalence 'Other surgical' (17) & 1.2 \% & 181 admissions \\
Prevalence 'Trauma surgical' (15) & 1.0 \% & 148 admissions \\
Prevalence 'Hematologic' (9) & 0.4 \% & 57 admissions \\
Prevalence 'Renal surgical' (16) & 0.2 \% & 27 admissions \\ 
\bottomrule
\end{tabular}}
\end{center}
\label{tab:appendix-resp-failure-statistics}
\end{table}
\FloatBarrier

\subsubsection*{Kidney function / Remaining-length-of-stay regression} \label{appsec:appendix-regressive-statistics}

\begin{figure}[h!]
\centering
\includegraphics[width=0.8\textwidth]{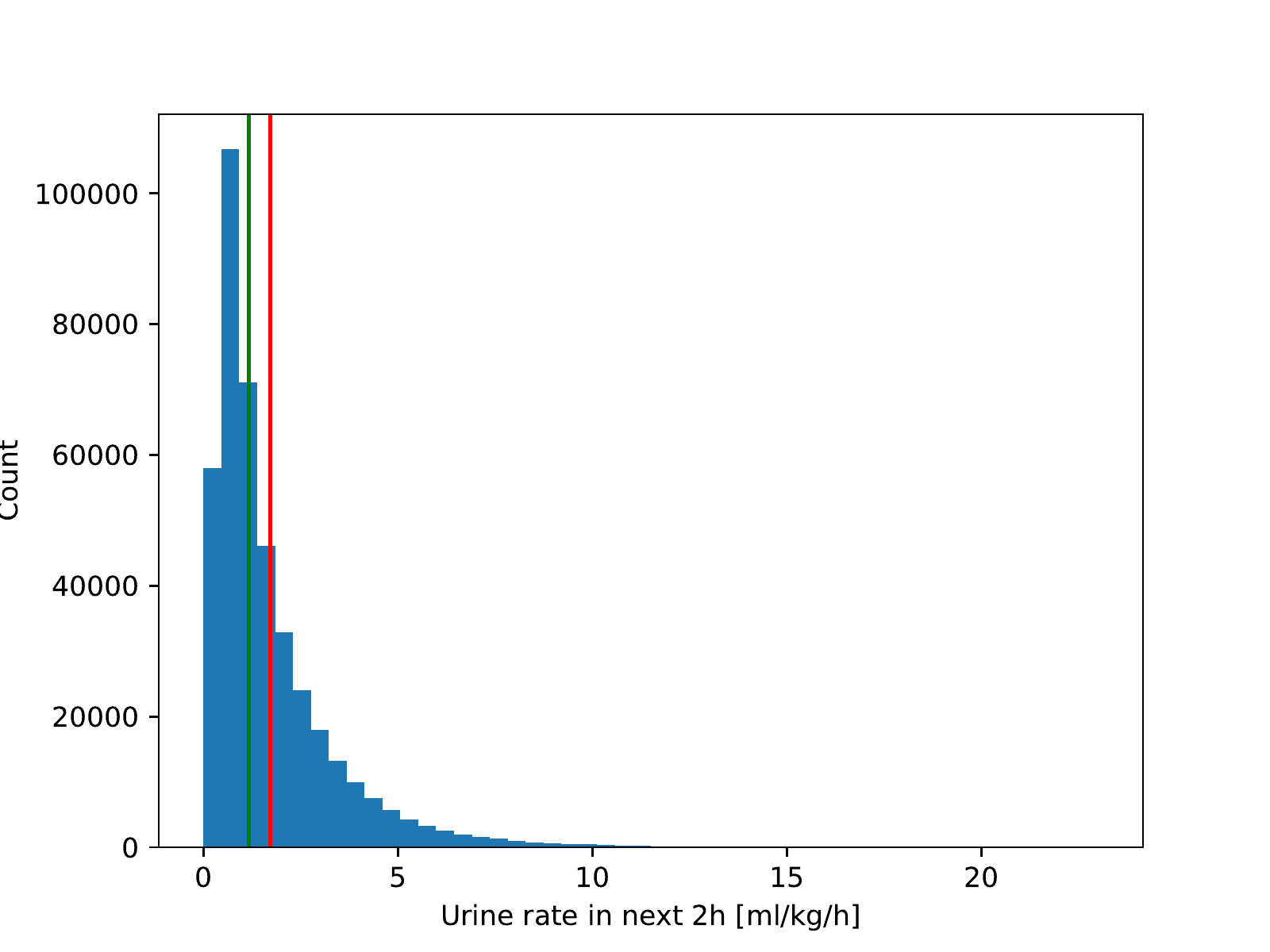}
\caption{Marginal distribution of the urine regression labels. The vertical red line denotes the mean, and the vertical
green line the median of the distribution.}
\label{fig:appendix-urine-label-dist}
\end{figure}
\FloatBarrier

\begin{figure}[h!]
\centering
\includegraphics[width=0.8\textwidth]{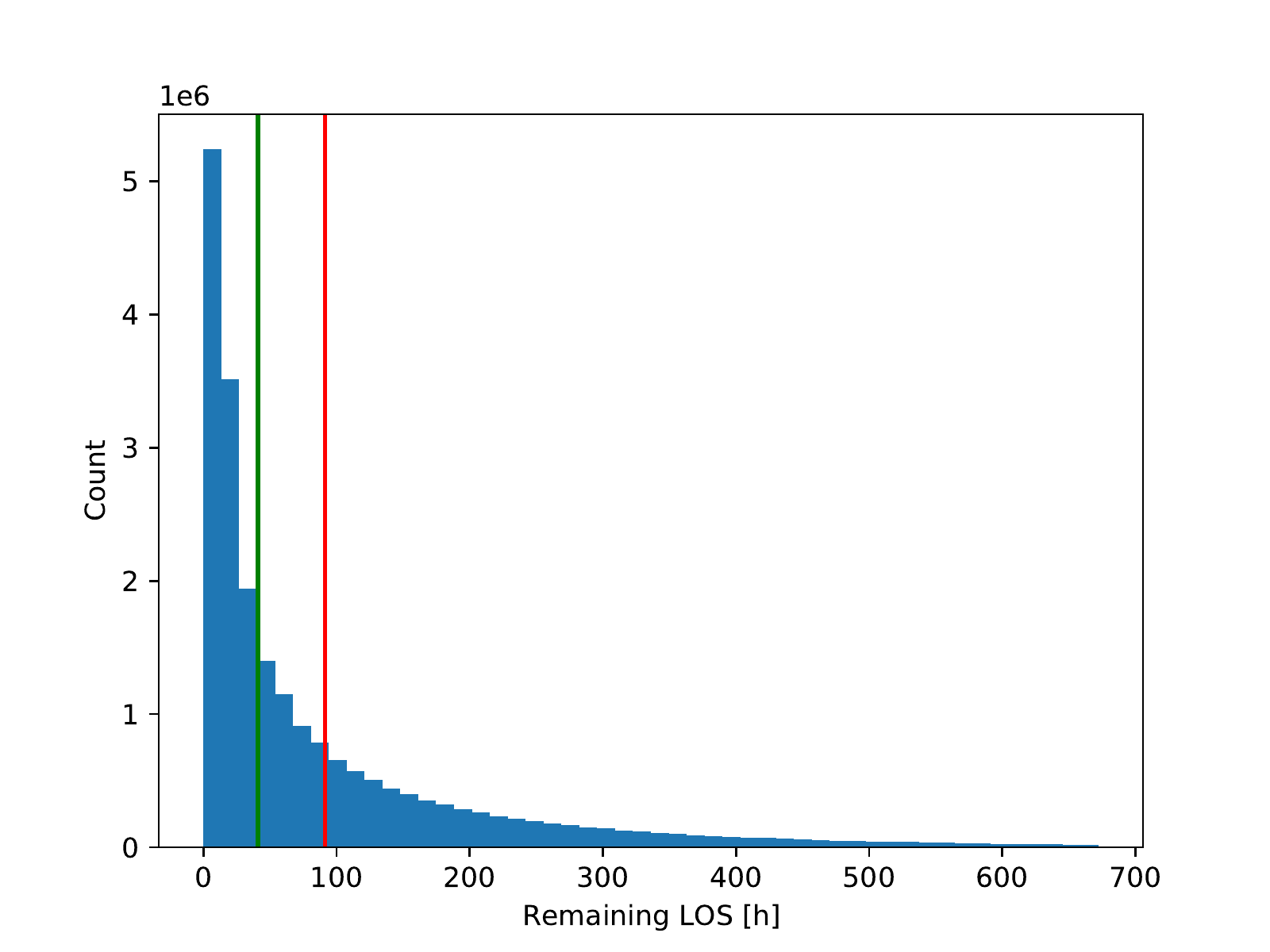}
\caption{Marginal distribution of the remaining length-of-stay regression labels. The vertical red line denotes the mean,
and the vertical green line the median of the distribution.}
\label{fig:appendix-los-label-dist}
\end{figure}
\FloatBarrier

\section{Appendix B: HiRID-ICU Pipeline Details}

In this section we provide a description of the full pipeline of the HiRID-ICU Benchmark and go through the main steps in detail. We also address the ML Reproducibility checklist questions\footnote{\href{https://arxiv.org/pdf/2003.12206.pdf}{ML Reproducibility checklist}}. 
\begin{enumerate}\addtocounter{enumi}{-1}
    \item \textit{Data loading}. Request access and then download the data from \url{https://physionet.org/content/hirid/1.1.1/} (see how to request access to the data in the README section of the Software Repository\footnote{\url{https://github.com/ratschlab/HIRID-ICU-Benchmark/}}.) The data is available under the \href{https://physionet.org/content/hirid/view-license/1.1.1/}{PhysioNet Credentialed Health Data License 1.5.0}.
    \item \textit{Data preprocessing}. The preprocessing step is described in detail in the \textsc{Appendix A: Dataset Details}. It is composed of several stages: merge stage, resample stage, feature extraction stage.
    \item \textit{Task implementation}. In this stage we construct the state annotations and machine learning labels for our predefined set of tasks.
    \item \textit{Model training and evaluation.} We select and train both ML and DL models on the training and validation sets. We evaluate model performance on the test set.
\end{enumerate}

The script \texttt{run.py} in the \texttt{icu\_benchmarks} folder unites three stages of the pipeline which can be used individually: Preprocessing (Section B.1), Task implementation (Section B.2), and Evaluation (Section B.3). Each stage has an associated help function. Below we describe each stage in detail.
\section*{Preprocessing}\label{prep_stage}
The method \texttt{preprocess} unites all necessary steps to generate the input for the Deep Learning (DL) and Machine Learning (ML) models:
\begin{enumerate}
    \item  \texttt{run\_merge\_step}~---~ Converts ICU data from table format to a matrix format that can be input to the machine learning model (see details in \textsc{Appendix A: Dataset Details}).
    \item  \texttt{run\_resample\_step}~--- Re-samples data to a regularly sampled 5min resolution.
    \item \texttt{run\_feature\_extraction\_step}~--- Hand-engineered feature extraction for classical machine learning models.
    \item \texttt{run\_build\_dl}~--- Data splitting and preprocessing specific to ML.
\end{enumerate}

Running these steps requires the following arguments: 
\begin{enumerate}
    \item {\texttt{hirid-data-root}}~--- Path to the unpacked parquet files containing the HiRID data, downloaded from Physionet.
    \item {\texttt{work-dir}}~--- Path to the working directory of the user.
    \item {\texttt{var-ref-path}}~--- Path to the file containing meta-data about variables necessary for our pre-processing pipeline.
    \item {\texttt{split-path}}~--- Path to the exact split of the data for model training and evaluation.
    \item {\texttt{nr-workers}}~--- Number of workers to use during training. It depends on the user hardware capacity.
\end{enumerate}

To run the whole preprocessing, the command below is used:
\begin{lstlisting}[language=bash]

icu-benchmarks preprocess --hirid-data-root [path to unpacked parquet files] \
                          --work-dir [output directory] \
                          --var-ref-path ./preprocessing/resources/varref.tsv \
                          --split-path ./preprocessing/resources/split.tsv \
                          --nr-workers 8
\end{lstlisting}

\section*{Task Implementation}
The second stage of the pipeline, consisting of extracting task labels, is constructed around the following three steps:
\begin{enumerate}
    \item \texttt{imputation\_for\_endpoints}~--- Imputes data for endpoint generation. 
    \item \texttt{generate\_endpoints}~--- Generates the endpoints i.e. verify if conditions for events are matched at each timestep.
    \item \texttt{generate\_labels}~---  Extract final labels from endpoints.
\end{enumerate}
As mentioned in the paper, we construct a set of 6 tasks, relevant for healthcare workers: 
\begin{enumerate}
\item \texttt{Mortality\_At24Hours}~--- Mortality prediction at 24h after admission in the ICU (further Mortality).
\item \texttt{Phenotyping\_APACHEGroup}~--- Patient APACHE group classification 24h after admission in the ICU (further Patient Phenotyping).
\item  \texttt{Dynamic\_CircFailure\_12Hours}~--- Continuous prediction of occurrence of circulatory failure in the next 12 hours (further Circulatory Failure).
\item \texttt{Dynamic\_RespFailure\_12Hour}~--- Continuous prediction of occurrence of respiratory failure in the next 12 hours (further Respiratory Failure).
\item \texttt{Remaining\_LOS\_Reg}~--- Continuous prediction of the remaining stay duration (further Remaining LOS).
\item \texttt{Dynamic\_UrineOutput\_2Hours\_Reg}~--- Continuous prediction of patient urine production in the next 2h (further Kidney Function).
\end{enumerate}

\section*{Model Training and Evaluation}
In the final part of the pipeline, we provide the user with a choice of model to train: 
\begin{enumerate}
    \item from the list of DL models: Transformer, LSTM, GRU and Temporal CNN. All these models use the class \texttt{DLWrapper} defining methods mechanics for deep learning models' training and evaluation. 
    \item from the list of the ML Models: Gradient Boosting method and Logistic Regression. In the same manner as DL models, these models have a \texttt{MLWrapper} class.
\end{enumerate}

All models are trained on the training set. The validation set is used for early stopping and model selection through a random search. After the training procedure, the model performance is evaluated on the test set.

To run the training, the following command is used (as an example, we provide here the command for GRU model training for the \texttt{Dynamic\_CircFailure\_12Hours} task):

\begin{lstlisting}[language=bash]
icu-benchmarks train -c configs/hirid/Classification/GRU.gin \
                     -l [path to logdir] \
                     -t Dynamic_CircFailure_12Hours\
                     -sd 1111 
\end{lstlisting}
\begin{itemize}
    \item Argument \texttt{-l} specifies the path to the directory where the trained model and meta-data will be stored.
    \item Argument \texttt{-t} specifies the selected task. 
    \item Argument \texttt{-sd} specifies the random seed.
    \item Argument \texttt{-c} specifies path to \texttt{gin-config} configuration file. This file should include the path to the data after preprocessing, the task, the model with all hyper-parameters. 
\end{itemize}

An example of the configuration file for the GRU model is provided below\footnote{See examples of configuration files for all the tasks and models in the \texttt{config} folder of the software repository}. 

\begin{lstlisting}[language=Python]
import gin.torch.external_configurables
import icu_benchmarks.models.wrappers
import icu_benchmarks.models.encoders
import icu_benchmarks.models.utils
import icu_benchmarks.data.loader


EMB = 231
LR = 3e-4
HIDDEN = 64
NUM_CLASSES = 2
DEPTH = 1
BS = 64
EPOCHS = 1000
TASK = 'Dynamic_CircFailure_12Hours'
RES = 1
RES_LAB = 1
MAXLEN = 2016
LOSS_WEIGHT = None

# Train params
train_common.model = @DLWrapper()
train_common.dataset_fn = @ICUVariableLengthDataset
train_common.data_path = [path to data] # TODO add by user
train_common.weight = train_common.do_test = True

DLWrapper.encoder = @GRU()
DLWrapper.loss = @cross_entropy
DLWrapper.optimizer_fn = @Adam
DLWrapper.train.epochs = DLWrapper.train.batch_size = DLWrapper.train.patience = 10
DLWrapper.train.min_delta = 1e-4


ICUVariableLengthLoaderTables.splits = ['train','test','val']
ICUVariableLengthLoaderTables.task = ICUVariableLengthLoaderTables.data_resampling = ICUVariableLengthLoaderTables.label_resampling = ICUVariableLengthDataset.maxlen = 

# Optimizer params
Adam.lr = Adam.weight_decay = 1e-6

# Encoder params
GRU.input_dim = GRU.hidden_dim = GRU.layer_dim = GRU.num_classes = \end{lstlisting}

We define all macros at the top of the file and use them to configure some classes and functions. Among them, we have two custom function for data loading:
\begin{itemize}
    \item \texttt{ICUVariableLengthLoaderTables}~--- Main Loader class allowing to sample patient from the data. \\
    \item \texttt{ICUVariableLengthDataset}~--- \texttt{pytorch} dataset wrapper to ship data to the GPU.
\end{itemize}

\section*{Technical Specifics for Reproducibility}\label{appsec:detail-training} 
\paragraph{Libraries} A full list of libraries and the version we used is provided in the \texttt{environment.yml} file. The most important ones are the following: pytorch 1.8.1, scikit-learn 0.24.1, ignite 0.4.4, CUDA 10.2.89, cudNN 7.6.5.

\paragraph{Infrastructure}
We follow all guidelines provided by \texttt{pytorch} documentation to ensure reproducibility of our results. However, reproducibility across devices is not ensured. Thus we provide here the characteristics of our infrastructure. We trained deep learning methods on a single \texttt{NVIDIA RTX2080Ti} with a \texttt{Xeon E5-2630v4} core. For other methods we trained models on either \texttt{Xeon E5-2697v4} cores or \texttt{Xeon Gold 6140} cores.

\paragraph{Method}
For the main experiment, 10 different random initializations were used for each model; For the ablation study, 5 were used. For all models, we tuned specific hyper-parameters using random search with 100 iterations for stay-level tasks and 50 iterations for the dynamic ones. Each random set of parameters was run with 3 different random initializations. Early stopping with 10 step patience on the loss was used as a stopping criterion. We then chose hyper-parameters on AUPRC, balanced accuracy, and MAE for respectively, binary, multi-class, and regression tasks.

\paragraph{Complexity of training}
Among the deep learning methods, transformer memory complexity, with regard to the sequence length, is quadratic. This has forced us to reduce both batch size and the number of parameters of this model for the dynamic tasks. Also, to ensure reproducibility, using deterministic algorithms particularly slows down TCN training. With our hardware, training deep learning methods takes less than 1h for stay-level tasks and less than 6h for dynamic ones on a single GPU.
To reduce RAM consumption we provide the possibility to load data only at inference time with parameter \texttt{on\_RAM} in our loader. In that case, training is slightly slower but requires only around 8GB of RAM.

On the other hand, ML methods are faster to train but more RAM-consuming. Training any model takes less than 4h on 4 CPUs. However, peak memory can exceed 100GB when using hand-engineered features.

\section*{Hyperparameters Search}\label{appsec:hyper-parameters}
In this section we detail the range of hyperparameters we searched over and the one we used for our experiments.
\subsection*{Common Hyperparameters}
For the training of DL methods, we used certain parameters across multiple tasks and architecture as reported in Table \ref{tab:fixed-hp-dl}.
For the ML methods, we fixed certain parameters as in the original HiRID paper. For LGBM, we set the bagging frequency to 1,
the number of leaves to $2^\textbf{depth}$,  and the minimum number of children per leaf to 1000. In the rest of the section, we report the hyperparameters we searched over in our experiments.

\begin{table}[tbh!]
    \centering
    \begin{tabular}{l|c|c|c|c}
        \toprule
        Models & Optimizer  & Weight Decay & Batch Size Online & Batch Size Stay Level  \\
        \midrule
        LSTM & Adam  & 1e-6 & 64  & 64 \\ 
        GRU & Adam  & 1e-6 & 64  & 64 \\
        TCN & Adam  & 1e-6 & 64   & 64\\ 
        Transformer & Adam  & 1e-6 & 8   & 16 \\
        \bottomrule
        \end{tabular}
    \caption{Fixed Hyperparameters for DL Methods }
    \label{tab:fixed-hp-dl}
\end{table}

\newpage
\subsection*{Deep Learning model Hyperparameters}
In this section, we detail the range of hyperparameters considered for LSTM, GRU, TCN and transformer models.
\paragraph{LSTM}
The range of hyperparameters considered for the LSTM Model can be found in Table \ref{tab:hp-search-lstm}.
\begingroup

\begin{table}[tbh!]
    \centering
    \footnotesize
    \setlength\tabcolsep{2pt}

\begin{tabular}{l|c|c|c|c|c}

\toprule
Task &      Learning Rate &  Drop-out & Depth & Hidden Dimension & Loss Weighting \\
\midrule
\midrule

Mortality & (1e-5, 3e-5, \textbf{1e-4}, 3e-4) &  (0.0, \textbf{0.1}, 0.2, 0.3, 0.4) &     (\textbf{1}, 2, 3) &    (32, 64, \textbf{128}, 256) & (\textbf{None}, Balanced) \\
Phenotyping & (1e-5, 3e-5, 1e-4, \textbf{3e-4}) &  (\textbf{0.0}, 0.1, 0.2, 0.3, 0.4) &     (\textbf{1}, 2, 3) &    (32, 64, 128, \textbf{256}) & (None, \textbf{Balanced}) \\
\midrule
\midrule

Circ. Failure & (1e-5, 3e-5, 1e-4, \textbf{3e-4}) &  (0.0, 0.1, 0.2, 0.3,  \textbf{0.4}) &    (1, \textbf{2}, 3) &    (32, 64, 128, \textbf{256}) & (\textbf{None}, Balanced) \\
Resp. Failure & (1e-5, 3e-5,\textbf{ 1e-4}, 3e-4) &  (0.0, 0.1, 0.2, 0.3, \textbf{0.4}) &     (1, \textbf{2}, 3) &    (32, 64, 128, \textbf{256}) & (\textbf{None}, Balanced) \\

\midrule
\midrule

Urine Output & (1e-5, 3e-5, 1e-4, \textbf{3e-4}) &  (0.0, 0.1, 0.2, \textbf{0.3}, 0.4) &     (1, 2, \textbf{3}) &    (32, 64, \textbf{128}, 256) & N.A \\
Rem. LOS & (1e-5, 3e-5, 1e-4, \textbf{3e-4}) &  (0.0, 0.1, \textbf{0.2}, 0.3, 0.4) &     (1, 2, \textbf{3}) &    (32, 64, 128, \textbf{256}) & N.A \\
\bottomrule
\end{tabular}
    \caption{Hyperparameter search range for LSTM. In \textbf{bold} are the parameters we selected using random search.}
    \label{tab:hp-search-lstm}
\end{table}
\endgroup

\paragraph{GRU}The range of hyperparameters considered for the GRU Model can be found in Table \ref{tab:hp-search-gru}.
\begin{table}[tbh!]
    \centering
    \footnotesize
    \setlength\tabcolsep{2pt}

\begin{tabular}{l|c|c|c|c|c}

\toprule
Task &      Learning Rate &  Drop-out & Depth & Hidden Dimension & Loss Weighting \\
\midrule
\midrule

Mortality & (1e-5, 3e-5, 1e-4, \textbf{3e-4}) &  (\textbf{0.0}, 0.1, 0.2, 0.3, 0.4) &     (1, \textbf{2}, 3) &    (32, \textbf{64}, 128, 256) & (\textbf{None}, Balanced) \\
Phenotyping & (\textbf{1e-5}, 3e-5, 1e-4, 3e-4) &  (0.0, 0.1, 0.2, 0.3, \textbf{0.4}) &     (\textbf{1}, 2, 3) &    (32, 64, 128, \textbf{256}) & (None, \textbf{Balanced}) \\
\midrule
\midrule

Circ. Failure & (1e-5, 3e-5, 1e-4, \textbf{3e-4}) &  (0.0, 0.1, \textbf{0.2}, 0.3, 0.4) &     (1, \textbf{2}, 3) &    (32, 64, 128, \textbf{256}) & (\textbf{None}, Balanced) \\
Resp.Failure & (1e-5, 3e-5,\textbf{ 1e-4}, 3e-4) &  (0.0, 0.1, 0.2, 0.3, \textbf{0.4}) &     (1, 2, \textbf{3}) &     (32, 64, 128, \textbf{256}) & (\textbf{None}, Balanced) \\

\midrule
\midrule

Urine Output & (1e-5, 3e-5, 1e-4, \textbf{3e-4}) &  (0.0, 0.1, 0.2, \textbf{0.3}, 0.4) &     (1, 2, \textbf{3}) &    (32, 64, 128, \textbf{256}) & N.A \\
Rem. LOS & (1e-5, 3e-5, 1e-4, \textbf{3e-4}) &  (0.0, 0.1, 0.2, \textbf{0.3}, 0.4) &     (1, \textbf{2}, 3) &    (32, 64, \textbf{128}, 256) & N.A \\
\bottomrule
\end{tabular}
    \caption{Hyperparameter search range for GRU. In \textbf{bold} are the parameters we selected using random search. }
    \label{tab:hp-search-gru}
\end{table}

\newpage
\paragraph{TCN}The range of hyperparameters considered for the TCN Model can be found in Table \ref{tab:hp-search-tcn}. Note that we do not consider any depth factor as it is fully determined by the kernel size and the sequence length. 
\begin{table}[tbh!]
    \centering
    \footnotesize
    \setlength\tabcolsep{2pt}

\begin{tabular}{l|c|c|c|c|c}

\toprule
Task & Learning Rate &  Drop-out  & Kernel & Hidden Dimension & Loss Weighting \\
\midrule
\midrule

Mortality & (1e-5, 3e-5, \textbf{1e-4}, 3e-4) &  (\textbf{0.0}, 0.1, 0.2, 0.3, 0.4) &    (2, \textbf{4}, 8, 16, 32) &    (32, 64, 128, \textbf{256}) & (\textbf{None}, Balanced) \\
Phenotyping  & (1e-5, 3e-5, \textbf{1e-4}, 3e-4) &  (0.0, 0.1, \textbf{0.2}, 0.3, 0.4) &  (2, 4, 8, 16, \textbf{32}) &    (32, 64, \textbf{128}, 256) &
(None,\textbf{Balanced}) \\
\midrule
\midrule

Circ. Failure & (1e-5, 3e-5, 1e-4, \textbf{3e-4}) &  (0.0, \textbf{0.1}, 0.2, 0.3, 0.4) &   (2, \textbf{4}, 8, 16, 32) &    (32, 64, 128, \textbf{256}) & (\textbf{None}, Balanced) \\
Resp. Failure & (1e-5, 3e-5, 1e-4, \textbf{3e-4}) &   (0.0, 0.1, 0.2, 0.3, \textbf{0.4})&    (2, 4, \textbf{8}, 16, 32) &    (32, \textbf{64}, 128, 256) & (\textbf{None}, Balanced) \\

\midrule
\midrule

Urine Output & (1e-5, 3e-5, 1e-4, \textbf{3e-4}) &  (0.0, 0.1, \textbf{0.2}, 0.3, 0.4) &   (2, 4, \textbf{8}, 16, 32) &    (32, 64, 128, \textbf{256}) & N.A \\
Rem. LOS & (1e-5, 3e-5, 1e-4, \textbf{3e-4}) &  (0.0, 0.1, 0.2, \textbf{0.3}, 0.4) &     (2, 4, 8, 16, \textbf{32}) &    (32, 64, \textbf{128}, 256) & N.A \\
\bottomrule
\end{tabular}
    \caption{Hyperparameter search range for TCN. In \textbf{bold} are the parameters we selected using random search. }
    \label{tab:hp-search-tcn}
\end{table}

\paragraph{Transformer} The range of hyperparameters considered for Transformer Model can be found in Table \ref{tab:hp-search-transformer} and Table \ref{tab:hp-search-transformer_2}. We considered smaller parameters for online tasks due to GPU memory limitations.
\begin{table}[tbh!]
    \centering
    \footnotesize
\begin{tabular}{l|c|c|c|c}

\toprule
Task & Learning Rate &  Attention Drop-out  & Nb. Heads & Depth  \\
\midrule
\midrule

Mortality & (\textbf{1e-5}, 3e-5, 1e-4, 3e-4) &  (0.0, \textbf{0.1}, 0.2, 0.3, 0.4) &    (1, 2, \textbf{4}, 8) &   (1, \textbf{2}, 3)  \\
Phenotyping & (1e-5, 3e-5, 1e-4, \textbf{3e-4}) &  (0.0, 0.1, \textbf{0.2}, 0.3, 0.4) &   (1, 2, 4, \textbf{8}) &  (1, \textbf{2}, 3) \\
\midrule
\midrule

Circ. Failure & (1e-5, \textbf{3e-5}, 1e-4, 3e-4) &  (\textbf{0.0}, 0.1, 0.2, 0.3, 0.4) & (\textbf{1}, 2, 4) & (1, 2, \textbf{3})  \\
Resp. Failure & (1e-5, 3e-5, 1e-4, \textbf{3e-4}) &  (\textbf{0.0}, 0.1, 0.2, 0.3, 0.4) & (\textbf{1}, 2, 4) & (1, \textbf{2}, 3)  \\

\midrule
\midrule

Urine Output & (1e-5, \textbf{3e-5}, 1e-4, 3e-4) &  (0.0, \textbf{0.1}, 0.2, 0.3, 0.4) &   (\textbf{1}, 2, 4, 8) & \textbf{1} \\
Rem. LOS & (1e-5, 3e-5, 1e-4, \textbf{3e-4}) &  (\textbf{0.0}, 0.1, 0.2, 0.3, 0.4) &    (\textbf{1}, 2, 4, 8)&\textbf{1} \\
\bottomrule
\end{tabular}
    \caption{Hyperparameter search range for the Transformer. In \textbf{bold} are the parameters we selected using random search. }
    \label{tab:hp-search-transformer}
\end{table}

\begin{table}[tbh!]
    \centering
    \footnotesize
\begin{tabular}{l|c|c|c}

\toprule
Task & Attention Drop-out & Hidden Dimension & Loss Weighting \\
\midrule
\midrule
Mortality & (\textbf{0.0}, 0.1, 0.2, 0.3, 0.4) &  (32, 64, 128, \textbf{256}) & (\textbf{None}, Balanced) \\
Phenotyping & (\textbf{0.0}, 0.1, 0.2, 0.3, 0.4) &  (\textbf{32}, 64, 128, 256) & (None, \textbf{Balanced}) \\
\midrule
\midrule
Circ. Failure & (0.0, 0.1, 0.2, 0.3, \textbf{0.4}) & (32, 64, \textbf{128}) & (\textbf{None}, Balanced) \\
Resp. Failure & (0.0, 0.1, 0.2, \textbf{0.3}, 0.4) &  (32, \textbf{64}, 128) & (\textbf{None}, Balanced) \\
\midrule
\midrule
Urine Output & (0.0, \textbf{0.1}, 0.2, 0.3, 0.4) &  (32, \textbf{64}, 128) & N.A \\
Rem. LOS  &  (\textbf{0.0}, 0.1, 0.2, 0.3, 0.4) & (32, 64, \textbf{128})  & N.A \\
\bottomrule
\end{tabular}
    \caption{Hyperparameter search range for Transformer. In \textbf{bold} are the parameters we selected using random search. }
    \label{tab:hp-search-transformer_2}
\end{table}

\newpage
\subsection*{Machine Learning Models Hyperparameters}
\paragraph{Gradient Boosting} The range of hyperparameters considered for the gradient boosting method, LightGBM framework\footnote{\url{https://lightgbm.readthedocs.io/en/latest/}} can be found in Table \ref{tab:hp-search-gb} and \ref{tab:hp-search-gb-feat} :
\begin{table}[tbh!]
    \centering
\begin{tabular}{l|c|c|c}

\toprule
Task & Depth &  Colsample\_bytree\tablefootnote{Subsample ratio of columns when constructing each tree.} & Subsample\tablefootnote{Subsample ratio of the training instance} \\
\midrule
\midrule
Mortality & (3, 4, 5, 6, \textbf{7})  & (0.33, 0.66, \textbf{1.00}) & (\textbf{0.33}, 0.66, 1.00) \\
Phenotyping & (\textbf{3}, 4, 5, 6, 7) & (0.33, \textbf{0.66}, 1.00)& (0.33, \textbf{0.66}, 1.00) \\
\midrule
\midrule
Circulatory Failure &(3, \textbf{4}, 5, 6, 7) & (\textbf{0.33}, 0.66, 1.00) & (\textbf{0.33}, 0.66, 1.00) \\
Respiratory Failure & (3, 4, 5, 6, \textbf{7}) & (\textbf{0.33}, 0.66, 1.00) & (\textbf{0.33}, 0.66, 1.00) \\
\midrule
\midrule
Urine Output & (3,\textbf{4}, 5, 6, 7) & (0.33, 0.66, \textbf{1.00})& (0.33, 0.66, \textbf{1.00}) \\
Remaining Length-of-Stay & (3, 4, 5, 6, \textbf{7}) & (\textbf{0.33}, 0.66, 1.00)& (0.33, 0.66, \textbf{1.00}) \\
\bottomrule
\end{tabular}
    \caption{Hyperparameter search range for LGBM. In \textbf{bold} are the parameters we selected using random search.}
    \label{tab:hp-search-gb}
\end{table}

\begin{table}[tbh!]
    \centering
\begin{tabular}{l|c|c|c}

\toprule
Task & Depth &  Colsample\_bytree\tablefootnote{Subsample ratio of columns when constructing each tree.} & Subsample\tablefootnote{Subsample ratio of the training instance} \\
\midrule
\midrule
Mortality & (3, 4, 5, 6, \textbf{7})  &(0.33, 0.66, \textbf{1.00}) & (0.33, 0.66, \textbf{1.00}) \\
Phenotyping & (3, 4, \textbf{5}, 6, 7) & (\textbf{0.33}, 0.66, 1.00)& (\textbf{0.33}, 0.66, 1.00) \\
\midrule
\midrule
Circulatory Failure & (3, \textbf{4}, 5, 6, 7) & (\textbf{0.33}, 0.66, 1.00) & (0.33, \textbf{0.66}, 1.00) \\
Respiratory Failure &(3, 4, 5, \textbf{6}, 7) & (\textbf{0.33}, 0.66, 1.00) & (0.33, \textbf{0.66}, 1.00) \\
\midrule
\midrule
Urine Output&(3, 4, 5, \textbf{6}, 7) &(0.33, 0.66, \textbf{1.00}) & (0.33, 0.66, \textbf{1.00}) \\
Remaining Length-of-Stay & (3, 4, 5, 6, \textbf{7}) & (0.33, \textbf{0.66}, 1.00)& (\textbf{0.33}, 0.66, 1.00) \\
\bottomrule
\end{tabular}
    \caption{Hyper-parameters range for LGBM w. features. In \textbf{bold} are the parameters we selected using random search.}
    \label{tab:hp-search-gb-feat}
\end{table}

\paragraph{Logistic Regression} 
The search range of hyperparameters considered for Logistic Regression\footnote{scikit-learn framework} can be found in Table \ref{tab:hp-search-lr}:
\begin{table}[tbh!]
    \centering
\begin{tabular}{l|c|c}

\toprule
Task & C &  Penalty \\
\midrule
\midrule
Mortality & (0.001, 0.01, 0.1, \textbf{1}, 10) & ('l1', \textbf{'l2'})   \\
Phenotyping &(0.001, 0.01, \textbf{0.1}, 1, 10) & ('l1',  \textbf{'l2'}) \\
\midrule
\midrule
Circulatory Failure & (0.001, \textbf{0.01}, 0.1, 1, 10) & ('l1',  \textbf{'l2'}) \\
Respiratory Failure &(\textbf{0.001}, 0.01, 0.1, 1, 10) &('l1', \textbf{'l2'})   \\
\bottomrule
\end{tabular}
    \caption{Hyperparameter search range for Logistic Regression. In \textbf{bold} are the parameters we selected using random search.}
    \label{tab:hp-search-lr}
\end{table}

\end{document}